\documentclass[runningheads]{llncs}

\usepackage{eccv}

\usepackage{eccvabbrv}

\usepackage{graphicx}
\usepackage{booktabs}

\usepackage[accsupp]{axessibility}  

\usepackage[pagebackref,breaklinks,colorlinks,citecolor=myblue]{hyperref} 

\usepackage{orcidlink}

\usepackage{microtype}
\usepackage{graphicx}
\usepackage{booktabs} 
\usepackage{rotating}
\usepackage{amssymb}
\usepackage{pifont}

\usepackage[table]{xcolor}
\usepackage{colortbl}
\usepackage{threeparttable}

\usepackage{makecell}   

\usepackage{amsmath}
\usepackage{amssymb}
\usepackage{mathtools}
\usepackage[capitalize,noabbrev]{cleveref}
\usepackage{xspace}
\usepackage{array}
\usepackage{multirow}
\usepackage[numbers,sort&compress]{natbib}
\usepackage{wrapfig}

\definecolor{myblue}{RGB}{30,144,255}
\newcommand{\our}{{Articulat3D}\xspace}

\newcommand{\new}[1]{{\color{blue}}}

\newcommand{\tf}[1]{\mathbf{#1}}

\newcommand{\gmean}{\boldsymbol{\mu}}

\newcommand{\motioncoef}{\mathbf{w}}

\newcommand{\ftf}[3]{{#1}_{#3 \rightarrow #2}}

\usepackage{epsfig}
\usepackage{xcolor}
\usepackage{graphicx}
\usepackage{float}
\usepackage[utf8]{inputenc}

\usepackage{caption}
\usepackage{subcaption}
\usepackage{array}
\usepackage{multirow}
\usepackage{etoolbox}
\usepackage{anyfontsize}
\usepackage{booktabs}
\usepackage{colortbl}
\usepackage{bm}
\usepackage{pifont}
\usepackage{adjustbox}
\usepackage[accsupp]{axessibility}

\newcommand{\cmark}{\textcolor{green}{\ding{51}}} 
\newcommand{\ccmark}{\cmark\kern-0.4em\cmark} 
\newcommand{\xmark}{\textcolor{red}{\ding{55}}} 

\newcolumntype{R}[2]{%
    >{\adjustbox{angle=#1,lap=\width-(#2)}\bgroup}%
    l%
    <{\egroup}%
}
\newcommand*\rot{\multicolumn{1}{R{35}{1em}}}

\begin{document}

\title{Articulat3D: Reconstructing Articulated Digital Twins From Monocular Videos with Geometric and Motion Constraints} 

\titlerunning{Articulat3D}

\author{Lijun Guo*\inst{1} \and
Haoyu Zhao*\inst{2, 3} \orcidlink{0009-0009-8274-2262} \and
Xingyue Zhao\inst{4} \and
Rong Fu \inst{5} \and
Linghao Zhuang\inst{1} \and
Siteng Huang\inst{6}\and
Zhongyu Li\inst{2, 3}\and
Hua Zou\textdagger\inst{1}
}

\authorrunning{L. Guo et al.}

\institute{
$^1$ Wuhan University, \quad
$^2$ Hong Kong Embodied AI Lab, \quad \\
$^3$ The Chinese University of Hong Kong, \quad
$^4$ Peking Union Medical College, \\
$^5$ University of Macau, \quad
$^6$ Zhejiang University
}
\maketitle

{
\renewcommand{\thefootnote}{\fnsymbol{footnote}}
\footnotetext{* Equal contributions.}
\footnotetext{\textdagger Corresponding Author.}
}

\begin{abstract}
Building high-fidelity digital twins of articulated objects from visual data remains a central challenge. Existing approaches depend on multi-view captures of the object in discrete, static states, which severely constrains their real-world scalability. 
In this paper, we introduce \our, a novel framework that constructs such digital twins from casually captured monocular videos by jointly enforcing explicit 3D geometric and motion constraints.
We first propose Motion Prior–Driven Initialization, which leverages 3D point tracks to exploit the low-dimensional structure of articulated motion. By modeling scene dynamics with a compact set of motion bases, we facilitate soft decomposition of the scene into multiple
rigidly-moving groups. Building on this initialization, we introduce Geometric and Motion Constraints Refinement, which enforces physically plausible articulation through learnable kinematic primitives parameterized by a joint axis, a pivot point, and per-frame motion scalars, yielding reconstructions that are both geometrically accurate and temporally coherent.
Extensive experiments demonstrate that \our achieves state-of-the-art performance on synthetic benchmarks and real-world casually captured monocular videos, significantly advancing the feasibility of digital twin creation under uncontrolled real-world conditions. Here is our \href{https://maxwell-zhao.github.io/Articulat3D/}{project page}.
\end{abstract}


\section{Introduction}
Articulated objects, prevalent in our daily life, are a major focus in computer vision and robotics~\cite{weng2024neural,luo2025physpart,liu2024cage,yang2023reconstructing,zhao2025towards}. Reconstructing an interactable digital twin of an articulated object—capturing its part geometry, visual appearance, and articulation parameters—enables a wide range of downstream applications, including augmented reality~\cite{zhao2025physsplat}, robotics simulation~\cite{zhao2025high}, and interactive scene understanding~\cite{liu2025videoartgs}. Among possible sensing modalities, monocular videos are particularly attractive because they are cheap to acquire at scale and readily available from casual capture and internet content, making it feasible for robotic agents to model objects directly from onboard observations. Achieving accurate articulated digital twins from such monocular inputs is therefore a key step toward scalable scene and object modeling and toward reducing the sim-to-real gap in robotic manipulation and interaction~\cite{torne2024reconciling,kerr2024robot}.

Recent approaches to reconstructing articulated objects can be broadly categorized into three families based on the way to estimate articulation parameters. 
One line of work uses feed-forward predictors to infer articulation directly from images or videos~\cite{mandi2024real2code,le2024articulate,jiang2022ditto}. While efficient at test time, these methods often depend on an external library for object retrieval or require data for fine-tuning, which limits scalability and robustness in real-world deployment. A second, more common family reconstructs objects by explicitly estimating joint parameters from multi-view images of the object in two or more discrete states~\cite{liu2023paris,liu2025building,weng2024neural}. While benefiting from geometric constraints, they necessitate controlled data capture setups and, crucially, require precise camera intrinsics and extrinsics for reconstruction, which is impractical for spontaneous, in-the-wild scenarios. A far more practical and scalable paradigm is reconstructing articulated objects from casually captured monocular videos which enables the ability to learn from internet videos and allows robotic agents to model objects directly from their visual observations~\cite{liu2025videoartgs,peng2025itaco,kerr2024robot}. While seemingly more scalable, these methods still implicitly rely on an initial static scan of the object, typically requiring the video to start with a segment where the camera moves around the object.

\begin{figure}[t]
  \centering
  \begin{minipage}[t]{0.49\textwidth} 
    \vspace{0pt}
    \centering
    \small
    \setlength{\tabcolsep}{2pt}
    \resizebox{\linewidth}{!}{
        \begin{tabular}{cccccc|l}
        \rot{no multi-state capture} & \rot{explicit articulation model} & \rot{physically-grounded geometry} & \rot{monocular video input} & \rot{no separate static scan} & \rot{joint optimization of frames} & \\
        \toprule
        \cmark & \xmark & \xmark & \cmark & \cmark & \xmark & Shape of Motion~\cite{wang2025shape}\\
        \cmark & \cmark & \xmark & \xmark & \cmark & \xmark & Articulate Anything~\cite{le2024articulate}\\
        \xmark & \cmark & \xmark & \xmark & \xmark & \xmark & PARIS~\cite{liu2023paris}\\
        \xmark & \cmark & \cmark & \xmark & \xmark & \xmark & ArtGS~\cite{liu2025building}\\
        \xmark & \cmark & \cmark & \xmark & \xmark & \xmark & ArticulatedGS~\cite{guo2025articulatedgs}\\
        \cmark & \cmark & \cmark & \xmark & \xmark & \xmark & iTACO~\cite{peng2025itaco}\\
        \cmark & \cmark & \cmark & \cmark & \xmark & \xmark & RSRD~\cite{kerr2024robot}\\
        \cmark & \cmark & \cmark & \cmark & \xmark & \xmark & VideoArtGS~\cite{liu2025videoartgs}\\
        \hline
        \rowcolor[HTML]{D7F6FF}\cmark & \cmark & \cmark & \cmark & \cmark & \cmark & \textbf{Ours}\\
        \end{tabular}
    }
    \captionof{table}{\textbf{Comparison to Concurrent Works.} Articulat3D uniquely enables the reconstruction of explicit articulation models from casual monocular videos by enforcing explicit physical motion and rigorous geometric joint parameter constraints.}
    \label{tab:comparison_concurrent}
  \end{minipage}
  \hfill 
  \begin{minipage}[t]{0.48\textwidth} 
    \vspace{15pt}
    \centering
    \includegraphics[width=\linewidth]{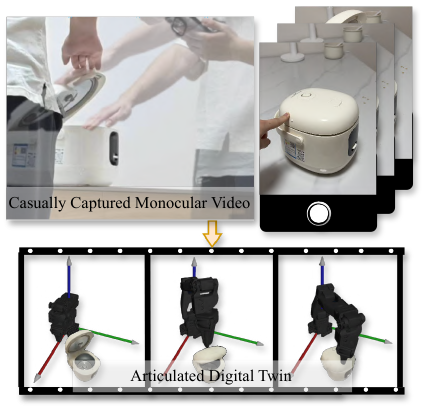}
    \vspace{-18pt}
    \caption{\textbf{From casual monocular video to interactive digital twin.} Given a video of an object's articulation, Articulat3D produces physically-consistent models that are ready for interaction and manipulation in simulation environments.}
    \label{fig:teasor}
  \end{minipage}
\end{figure}

More fundamentally, a pervasive but overlooked limitation shared by these paradigms is their reliance on optimizing motion parameters (\eg, dual quaternions) for discrete states or individual frames independently. This frame-wise optimization is inherently disjointed; the model focuses on fitting the visual appearance at specific timestamps rather than learning the actual evolutionary process of the motion. In reality, articulated motion, such as a door transitioning from closed to open, follows a continuous physical trajectory. By treating each frame as an isolated optimization target, existing methods fail to capture this temporal coherence, leading to reconstructions that lack physical plausibility during the transition phases, which limits their use on in-the-wild videos that start with immediate motion. As summarized in Tab.~\ref{tab:comparison_concurrent}, existing methods often lack the ability to jointly optimize all frames under explicit articulation constraints without a separate static scan.
We therefore ask this question: \textit{how can we 
reconstruct articulated digital twins from just casually captured monocular videos by imbuing the model with an understanding of the fundamental geometric constraints that govern articulated motion}?

To this end, we present \textbf{\our}, a novel framework for reconstructing high-fidelity, interactable articulated digital twins from a single monocular video by modeling motion as a \emph{continuous, constrained trajectory} rather than a collection of independent per-frame states, as shown in Fig.~\ref{fig:teasor}. 
\our follows a two-stage optimization strategy. First, in \emph{Motion Prior-Driven Initialization}, instead of optimizing per-frame displacements, we exploit the low-dimensional structure of articulated movement by modeling the scene’s dynamics through a compact set of motion bases. By representing each trajectory as a linear combination of these bases, we effectively regularize the motion space, ensuring global temporal consistency from the outset.
Second, in \emph{Geometric and Motion Constraints Refinement}, we explicitly parameterize articulation with learnable kinematic primitives—joint axes, pivot points, and per-frame motion scalars—and enforce these constraints in a differentiable optimization loop, encouraging physically plausible motion throughout the entire sequence, including transition phases.
Extensive experiments on both synthetic benchmarks and real-world casual captures demonstrate that our method significantly outperforms existing approaches, achieving state-of-the-art accuracy in both geometry reconstruction and motion estimation.
In summary, our work makes the following contributions:
\begin{itemize}
    \item We present \textbf{\our}, a novel framework for high-fidelity, interactable digital twin reconstruction from monocular videos.
    \item We introduce a two-stage optimization strategy: Motion Prior-Driven Initialization exploits the low-dimensional structure of 3D point tracks via motion bases, and Geometric and Motion Constraints Refinement enforces strict physical plausibility through learnable kinematic primitives.
    \item We introduce two challenging benchmarks featuring multi-part kinematics and uncontrolled captures. Experiments demonstrate that our method achieves state-of-the-art performance in all metrics.
\end{itemize}

\section{Related Work}
\subsection{Dynamic Scene Reconstruction}
Dynamic scene reconstruction is a long-standing challenge. One significant line of research focuses on jointly estimating camera poses and scene geometry. Pioneering optimization-based methods~\cite{zhang2025tapip3d} like DROID-SLAM~\cite{teed2021droid}, and Mega-SaM~\cite{li2025megasam} established robust frameworks for this task. More recently, foundation models have provided a powerful basis for 3D reconstruction, with DUSt3R~\cite{wang2024dust3r} being a prominent example. Subsequent works extend these models to better handle dynamic content, such as MonST3R~\cite{zhang2024monst3r}, which fine-tunes DUSt3R on dynamic datasets, and CUT3R~\cite{wang2025continuous}, which proposes a continuous model for online reconstruction. While these methods effectively recover camera parameters and static geometry, modeling scene dynamics remains a formidable challenge.

A parallel and highly successful approach represents dynamic scenes with a deformation field, evolving from signed distance functions (SDFs) to modern neural representations. In particular, 4D extensions of 3D Gaussian Splatting~\cite{duan20244d,zhao2025physsplat,luiten2024dynamic,wang2025shape,wu20244d,yang2023real,zhao2024hfgs,zhao2024sg,zhao20243d} achieve state-of-the-art results in shape reconstruction and novel-view rendering. However, a critical limitation is that implicit deformation field is difficult to convert into an explicit motion representation, such as a rigid transformation matrix or an articulated structure. While concurrent works like Shape of Motion~\cite{wang2025shape} attempt to extract motion from gaussians, they still treat motion as generic displacements rather than constrained articulated trajectories, a gap our work specifically addresses. \our integrates an explicit, constraint-driven articulation model into the Gaussian framework. By regularizing motion with kinematic primitives, our method produces a digital twin that is both visually accurate and physically controllable for downstream simulation across diverse articulated-object configurations and interaction scenarios.

\subsection{Articulated Object Reconstruction}
Reconstruction of articulated objects is a long-standing task in computer vision and robotics, which presents a dual challenge: one must solve for both the part-level geometry and the underlying articulation parameters. One family of methods employs large language and multi-modal models to predict both part segmentation and joint parameters~\cite{heppert2023carto,wei2022self,mandi2024real2code,goyal2025geopard,xia2025drawer}, while some similar methods only predict articulation
parameters~\cite{hu2017learning,yi2018deep}. Leveraging vision-language models, Articulate Anything~\cite{le2024articulate} take image or video as input to generate a code representation for articulated objects. Their fundamental limitation, however, is a reliance on large, existing mesh libraries, which prevents them from generalizing to unseen object categories in real-world setting.

Another line of work attempts to infer joint parameters and build part-level geometry of articulated objects simultaneously~\cite{liu2025building,jiang2022ditto,liu2023paris,weng2024neural,guo2025articulatedgs,wu2025reartgs,wu2025reartgs++}. They rely on multi-view observations at discrete multi-state. These methods leverage strong geometric constraints, which simplify the problem but require impractical and controlled data capture setups. For example, ArtGS~\cite{liu2025building} and ArticulatedGS~\cite{guo2025articulatedgs} must be provided with two static 3D reconstructions of the object at different articulation states and assume all observations are aligned to the same coordinates. While this simplifies the correspondence problem, the underlying data acquisition process remains impractical for real-world use.

A more practical but far more challenging setting is reconstruction from just a monocular video. Existing methods are typically limited to simple objects~\cite{song2024reacto,peng2025itaco,zhou2025monomobility} or implicitly rely on an initial static scan of the object~\cite{kerr2024robot,peng2025itaco,liu2025videoartgs}. For example, RSRD~\cite{kerr2024robot} and iTACO~\cite{peng2025itaco} require a pre-scanned 3D model before analyzing the dynamic video. Even methods designed for monocular input, such as VideoArtGS~\cite{liu2025videoartgs}, still needs an initial static sequence for geometric initialization. This prerequisite fundamentally limits their applicability to in-the-wild videos that often begin with immediate motion. In contrast, our \our bypasses the need for static scans or templates by leveraging low-dimensional motion bases and explicit kinematic constraints, enabling robust joint optimization of the entire motion trajectory from the very first frame.

\section{Our Methodology}

\begin{figure*}[t]
  \centering
    \includegraphics[width=\linewidth]{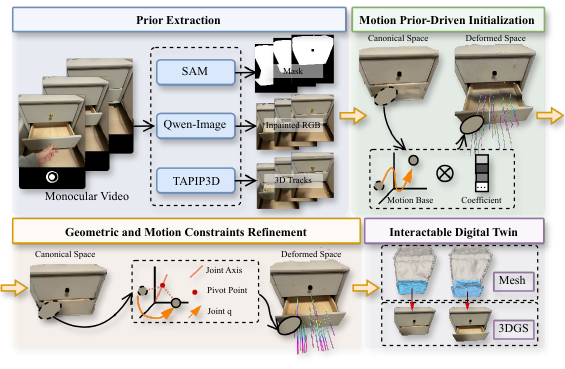}\\
  \vspace{-10pt}
  \caption{\textbf{Overview of \our \raisebox{-0.15\baselineskip}{\includegraphics[scale=0.03]{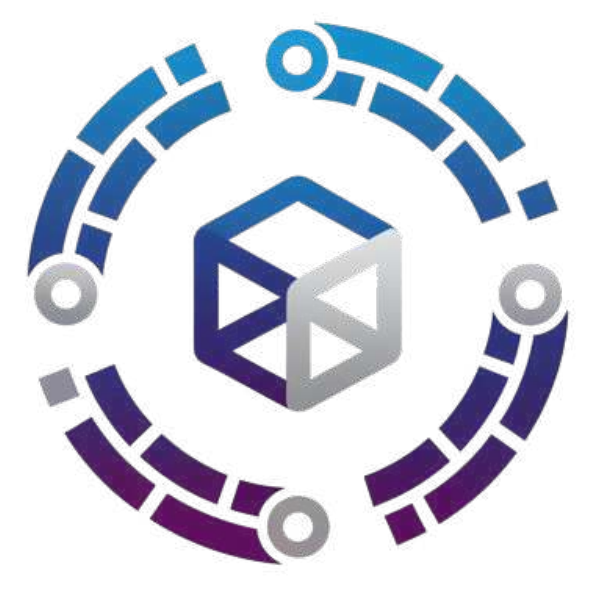}}}. Given a monocular video, \our reconstructs a high-fidelity articulated digital twin via two stages. (1) \textit{Motion Prior–Driven Initialization} exploits the low-dimensional structure of 3D point tracks by modeling them as a linear combination of shared motion bases. (2) \textit{Geometric and Motion Constraints Refinement} then enforces physical plausibility by optimizing learnable kinematic primitives, including joint axes and pivot points.  This joint optimization ensures that the final reconstruction is both geometrically accurate and temporally coherent.}
  \label{fig:pipeline}
\end{figure*}

As illustrated in Fig.~\ref{fig:pipeline}, \textbf{\our} reconstructs a high-fidelity digital twin from a casual monocular video sequence $X = \{x_i\}_{i=1}^N$ through a two-stage optimization pipeline. We first preprocess the input frames by segmenting the object using SAM3~\cite{carion2025sam}, filtering occluding hands via Qwen-Image~\cite{wu2025qwen}, and extracting initial 3D motion trajectories with TAPIP3D~\cite{zhang2025tapip3d} to serve as a foundational motion prior. In the first stage, we introduce \textbf{Motion Prior-Driven Initialization} to exploit the low-dimensional structure of articulated motion by modeling scene dynamics with a compact set of motion bases, where each noisy trajectory is fused into a coherent 3D Gaussian Splatting (3DGS) representation as a linear combination of these bases. Building upon this, the second stage, \textbf{Geometric and Motion Constraints Refinement}, enforces physical plausibility by introducing learnable kinematic primitives. By parameterizing component movements as either \textit{revolute} or \textit{prismatic} joints—defined by a joint axis, a pivot point, and per-frame motion scalars—we constrain each part to adhere strictly to rigid body transformations. This joint enforcement of geometric and kinematic constraints yields a reconstructed digital twin that is not only visually accurate and temporally coherent but also inherently controllable for downstream physical simulations.


\subsection{Motion Prior-Driven Initialization}
\label{sec:prior_init}

\noindent
\textbf{Prior Extraction.}
Casual captures of articulated objects often suffer from significant hand occlusions, leading to incomplete visual observations and distorted 3D reconstructions. To address this, we first employ Qwen-Image~\cite{wu2025qwen} to perform inpainting, removing occluding hands and recovering the holistic appearance of the object. This process yields a sequence of de-occluded images with corresponding complete segmentation masks $M = \{m_i\}^N_{i=1}$ generated by SAM3~\cite{carion2025sam}.
We also utilize TAPIP3D~\cite{zhang2025tapip3d} to obtain sparse 3D point trajectories with visibility. Each trajectory is independently classified as static, prismatic, revolute, or invalid by fitting simple geometric motion primitives. Static points are detected by bounded displacement around their visible mean. Prismatic candidates are obtained by robust 3D line fitting, while revolute candidates are obtained by fitting a plane and a circle to the projected trajectory, yielding an axis direction and pivot. We then cluster trajectories within each motion type using KMeans over primitive-aware features, including initial position, mean position, motion direction, pivot for revolute motion, and temporal displacement or angular signatures. The number of prismatic and revolute clusters is determined by the predicted joint-type prior. This produces a per-track part label, where static points share label 0 and each dynamic cluster corresponds to one candidate movable part.

\noindent 
\textbf{Motion Bases-based Modeling.} 
Directly optimizing per-Gaussian trajectories from noisy sparse tracks is ill-posed. Instead, we leverage the fact that articulated motion resides in a low-dimensional subspace. Inspired by Shape of Motion ~\cite{wang2025shape}, we model the scene dynamics using a compact set of $B$ shared, learnable $\mathbb{SE}(3)$ motion bases. 
Unlike Shape of Motion, which represents object motion with generic free motion bases and soft per-Gaussian mixture weights, our motion prior-driven initialization is designed to produce a part-aware motion representation from the beginning. Instead of allowing each Gaussian to freely combine multiple redundant bases, we bias the optimization toward compact, sparse, and near-discrete motion assignments that align with articulated parts. We also decouple the true articulation interval from static camera-orbiting frames, so temporal regularization is applied only over physically meaningful motion segments. As a result, motion prior-driven initialization provides a more coherent and interpretable part-level motion initialization, making the subsequent joint-axis and joint-value estimation more stable.

Given the TAPIP3D trajectories from the prior extraction step, we cluster them into $B$ part-level motion groups using spatiotemporal K-means (see Supp. Mat. for details), where each recovered cluster corresponds to one motion basis, including an identity basis for the static part. For each dynamic cluster $b$, we initialize its per-frame SE(3) basis $\mathbf{T}_{0 \rightarrow t}^{(b)}$ by solving a weighted Procrustes alignment between the canonical 3D tracks at frame $t_0$ and their observations at each target frame $t$. Each Gaussian is then assigned a strong one-hot motion coefficient according to its associated trajectory cluster. These coefficients remain trainable during Stage-I optimization, but are regularized to stay sparse and near-discrete, encouraging compact part-level motion instead of arbitrary soft combinations of redundant bases.
Specifically, for a 3D Gaussian at time $t$, its position $\gmean_t$ and orientation $\tf R_t$ are transformed from the canonical state $(\gmean_0, \tf R_0)$ via a rigid transformation $\ftf{\tf T}{t}{0} = [\ftf{\tf R}{t}{0} \mid \ftf{\tf t}{t}{0}] \in \mathbb{SE}(3)$:
\begin{equation}
    \gmean_t = \ftf{\tf R}{t}{0} \tf \mu_0 + \ftf{\tf t}{t}{0},\quad \tf {R}_t = \ftf{\tf R}{t}{0} \tf{R}_0.
\end{equation}

To represent complex articulated motion, each Gaussian is assigned a set of learnable coefficients $\{\motioncoef^{(b)}\}_{b=1}^B$, normalized via a softmax function such that $\sum_{b} \motioncoef^{(b)} = 1$. The aggregate transformation $\mathbf{T}_{0 \rightarrow t}$ is formulated as a weighted combination of the $B$ shared bases:
\begin{equation}
    \mathbf{T}_{0 \rightarrow t} = \text{Normalize} \left( \sum_{b=1}^B \motioncoef^{(b)}\ \mathbf{T}_{0 \rightarrow t}^{(b)} \right),
\end{equation}
where the normalization ensures the resulting transformation remains a valid $\mathbb{SE}(3)$ member (e.g., via SVD-based orthogonalization of the rotation component). This formulation regularizes the motion space, forcing Gaussians with similar semantic properties to share consistent, low-dimensional trajectories, thereby ensuring spatio-temporal coherence even in regions with sparse tracking data.


\subsection{Geometric and Motion Constraints Refinement}
\label{sec:geo_motion_refine}
While the motion bases in Sec.~\ref{sec:prior_init} provide a flexible initialization, they lack the physical rigour required for true digital twins, as they do not guarantee axis-aligned rigid-body motions. To enforce physical plausibility, we transition from unconstrained $\mathbb{SE}(3)$ bases to explicit kinematic primitives, specifically \textit{revolute} and \textit{prismatic} joints forcing each component to adhere to strict kinematic laws.

\noindent
\textbf{Kinematic Parameterization.}
We parameterize the motion of each object part $k$ using a normalized joint axis $\mathbf{a}_k \in \mathbb{S}^2$, a pivot point $\mathbf{c}_k \in \mathbb{R}^3$, and a per-frame scalar $q_k(t)$ representing displacement or rotation. These parameters define a per-part rigid transformation $\mathbf{T}_k(t) = [\mathbf{R}_k(t) \mid \mathbf{t}_k(t)] \in \mathbb{SE}(3)$.

For a \textbf{revolute joint} with rotation angle $\theta_k(t) = q_k(t)$, the transformation follows Rodrigues' formula around pivot $\mathbf{c}_k$:
\begin{equation}
    \mathbf{R}_k(t) = \exp(\theta_k(t)[\mathbf{a}_k]_\times), \quad \mathbf{t}_k(t) = (\mathbf{I} - \mathbf{R}_k(t))\mathbf{c}_k,
    \label{eq:revolute_transform}
\end{equation}
where $[\mathbf{a}_k]_\times$ denotes the skew-symmetric matrix of $\mathbf{a}_k$. For a \textbf{prismatic joint} with displacement $d_k(t) = q_k(t)$, the motion is a pure translation along the axis:
\begin{equation}
    \mathbf{R}_k(t) = \mathbf{I}, \quad \mathbf{t}_k(t) = d_k(t)\mathbf{a}_k.
    \label{eq:prismatic_transform}
\end{equation}
We provide more details about kinematic parameterization in our Supp. Mat.

\noindent
\textbf{Robust Joint Initialization}
To bridge the gap between soft bases and rigid joints, we first perform a hard assignment where each Gaussian $i$ is assigned to part $k = \arg\max_b w_i^{(b)}$ based on its learned coefficients. We then initialize the kinematic parameters by analyzing the aggregate motion:
1) \textit{Robust Trajectory Synthesis}: For each part $k$, we compute a robust center trajectory $\mathbf{c}_k(t)$ using a trimmed mean of the positions of its assigned Gaussians to filter out tracking noise.
2) \textit{Kinematic Discovery via PCA}: We apply Principal Component Analysis (PCA) to the centered trajectory $\{\mathbf{c}_k(t) - \bar{\mathbf{c}}_k\}$. The joint type is determined by the eigenvalue distribution: a single dominant eigenvalue indicates a \textbf{prismatic} joint (linear motion), while two significant eigenvalues suggest a \textbf{revolute} joint (planar arc). In the latter case, the joint axis $\mathbf{a}_k$ is initialized as the normal to the motion plane (the third principal component).
We provide more details about robust joint initialization in our Supp. Mat.

\begin{wrapfigure}{r}{0.45\textwidth}
    \centering
    \includegraphics[width=\linewidth]{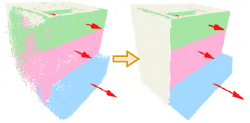}
      \vspace{-15pt}
    \caption{\textbf{Joint Refinement of Part Assignment.} Kinematic consistency corrects initial misalignments (left) into physically-grounded segments (right).}
    \vspace{-10pt}
    \label{fig:joint-refine}
\end{wrapfigure}

\noindent
\textbf{Joint Refinement of Part Assignment.}
Initial assignments based on motion bases may be noisy near kinematic boundaries. We refine these assignments jointly with the joint parameters by associating each Gaussian with a learnable latent vector $\mathbf{z}_i \in \mathbb{R}^K$ (initialized from $\mathbf{w}_i$), as shown in Fig.~\ref{fig:joint-refine}. Soft assignment probabilities $p_{i,k}$ are obtained via a softmax with temperature $\tau$.

To maintain strict rigidity in the forward pass, we apply the transformation of the most probable joint: $\mathbf{T}_i = \mathbf{T}_{k^*}$ where $k^* = \arg\max_k p_{i,k}$. For end-to-end optimization, we utilize a Straight-Through Estimator (STE) in the backward pass. The gradient of the loss $\mathcal{L}$ with respect to $\mathbf{z}_i$ is approximated by treating the transformation as a soft mixture $\sum_j p_{i,j}\mathbf{T}_j$:
\begin{equation}
    \frac{\partial \mathcal{L}}{\partial z_{i,k}} \approx p_{i,k} \sum_{j=1}^K \left\langle \frac{\partial \mathcal{L}}{\partial \mathbf{T}_i}, \mathbf{T}_j \right\rangle (\delta_{jk} - p_{i,j}),
    \label{eq:ste_grad}
\end{equation}
where $\delta_{jk}$ is the Kronecker delta and $\langle \cdot, \cdot \rangle$ denotes the Frobenius inner product. This "re-skinning" mechanism allows model to correct misassignments, ensuring each Gaussian aligns with the most physically consistent kinematic structure.

\subsection{Optimization}
\label{sec:optimization}

To reconstruct high-fidelity geometry and physically plausible motion, we jointly optimize the canonical Gaussian parameters $\mathcal{G}^c$, the latent assignments $\mathbf{z}_i$, and the kinematic articulation parameters $\Theta = \{\mathbf{a}_k, \mathbf{c}_k, q_k(t)\}$. The total objective function is defined as:
\begin{equation}
    \mathcal{L} = \mathcal{L}_{\text{render}} + \lambda_{\text{acc}} \mathcal{L}_{\text{acc}} + \lambda_{z} \mathcal{L}_{z}
\end{equation}
where $\lambda_{\text{acc}}$ and $\lambda_{z}$ are hyper-parameters balancing rendering accuracy, motion smoothness, and assignment consistency.

\noindent\textbf{Rendering Loss.} Following ArtGS~\cite{liu2025building}, $\mathcal{L}_{\text{render}}$ combines photometric accuracy with geometric supervision. Specifically, $\mathcal{L}_{\text{render}} = (1-\lambda_{\text{SSIM}})\mathcal{L}_1 + \lambda_{\text{SSIM}}\mathcal{L}_{\text{D-SSIM}} + \lambda_D \mathcal{L}_D$. For the depth supervision, we utilize monocular depth estimates $\bar{\bm{D}}$ (e.g., from~\cite{li2025megasam}) and apply a robust scale-invariant loss: $\mathcal{L}_D = \log(1 + \|\bm{D} - \bar{\bm{D}}\|_1)$, where $\bm{D}$ is the rendered depth map. To ensure temporal smoothness and mitigate monocular depth ambiguities, we employ an Acceleration Loss ($\mathcal{L}_{acc}$) and a Depth-Stability Loss ($\mathcal{L}_{z}$). A detailed mathematical formulation and a discussion of their physical significance are provided in our Supp. Mat.

\begin{table*}[t]
 \centering
    \caption{Comparison of \our with state-of-the-art baselines. We evaluate kinematic accuracy, reconstruction fidelity, 3D tracking, and view synthesis across synthetic and real-world datasets.}
    \vspace{-5pt}
    \label{tab:exp}
    \scriptsize
    \setlength{\tabcolsep}{2pt}
    \resizebox{\textwidth}{!}{%
        \begin{tabular}{cccccccccc}
        \toprule
        \multirow{2}{*}{Method} & \multicolumn{2}{c}{Joint Estimation} & \multicolumn{3}{c}{Reconstruction} & \multicolumn{1}{c}{Tracking} & \multicolumn{3}{c}{View Synthesis} \\
        \cmidrule(lr){2-3} \cmidrule(lr){4-6} \cmidrule(lr){7-7} \cmidrule(lr){8-10}
        & Axis Err $\downarrow$ & Pos Err $\downarrow$ & CD-w $\downarrow$ & CD-m $\downarrow$ & CD-s $\downarrow$ & EPE $\downarrow$ & PSNR $\uparrow$ & SSIM $\uparrow$ & LPIPS $\downarrow$ \\
        \midrule
    
        \multicolumn{10}{c}{\textbf{Video2Articulation-S}} \\
        \rowcolor{gray!10}RSRD~\citep{kerr2024robot} & 68.49  & 203.00  & 339.00 & 82.00 & 14.00 & N/A & 24.78 & 0.77 & 0.16 \\
        Articulate Anything~\citep{le2024articulate} & 49.85 & 81.00 & 11.00 & 59.00 & 7.00 & N/A & N/A & N/A & N/A \\
        \rowcolor{gray!10}iTACO~\citep{peng2025itaco} & 16.05 & 13.00 & 1.00 & 13.00 & 6.00 & N/A & N/A & N/A & N/A \\
        Shape of Motion~\cite{wang2025shape} & N/A & N/A & N/A & N/A & N/A & N/A & 31.88 & 0.96 & 0.05 \\
        \rowcolor[HTML]{D7F6FF} \textbf{\our} & \textbf{1.60} & \textbf{1.83}  & \textbf{0.82} & \textbf{1.12} & \textbf{1.82} & N/A & \textbf{35.91} & \textbf{0.98} & \textbf{0.05} \\
    
        \midrule[\heavyrulewidth]
    
        \multicolumn{10}{c}{\textbf{\our-Sim}} \\
        \rowcolor{gray!10} RSRD~\citep{kerr2024robot} & 55.75 & 91.52 & 69.16 & 55.41 & 10.05 & 3.14 & 28.19 & 0.94 & 0.06 \\
        Articulate Anything~\citep{le2024articulate} & 44.65 & 129.00 & 16.10 & 17.74 & 16.36 & 2.02 & N/A & N/A & N/A \\
        \rowcolor{gray!10} iTACO~\citep{peng2025itaco} & 48.50 & 38.00 & 3.13 & 1.63 & 3.72 & 0.14 & N/A & N/A & N/A \\
        Shape of Motion~\cite{wang2025shape} & N/A & N/A & N/A & N/A & N/A & 0.08 & 33.37 & 0.97 & 0.04 \\
        \rowcolor[HTML]{D7F6FF} \textbf{\our} & \textbf{0.53} & \textbf{0.65} & \textbf{0.76} & \textbf{0.84} & \textbf{0.85} & \textbf{0.04} & \textbf{37.80} & \textbf{0.98} & \textbf{0.03} \\
    
        \midrule[\heavyrulewidth]
        
        \multicolumn{10}{c}{\textbf{\our-Real}} \\
        \rowcolor{gray!10} RSRD~\citep{kerr2024robot} & N/A & N/A & N/A & N/A & N/A & N/A & 16.12 & 0.78 & 0.34 \\
        Shape of Motion~\cite{wang2025shape} & N/A & N/A & N/A & N/A & N/A & N/A & 24.13 & 0.91 & 0.15 \\
        \rowcolor[HTML]{D7F6FF}\textbf{\our} & N/A & N/A & N/A & N/A & N/A & N/A & \textbf{26.73} & \textbf{0.93} & \textbf{0.14} \\
        \bottomrule
        \end{tabular}
    }
\end{table*}

\begin{figure}[t]
    \centering
    \includegraphics[width=1.0\linewidth]{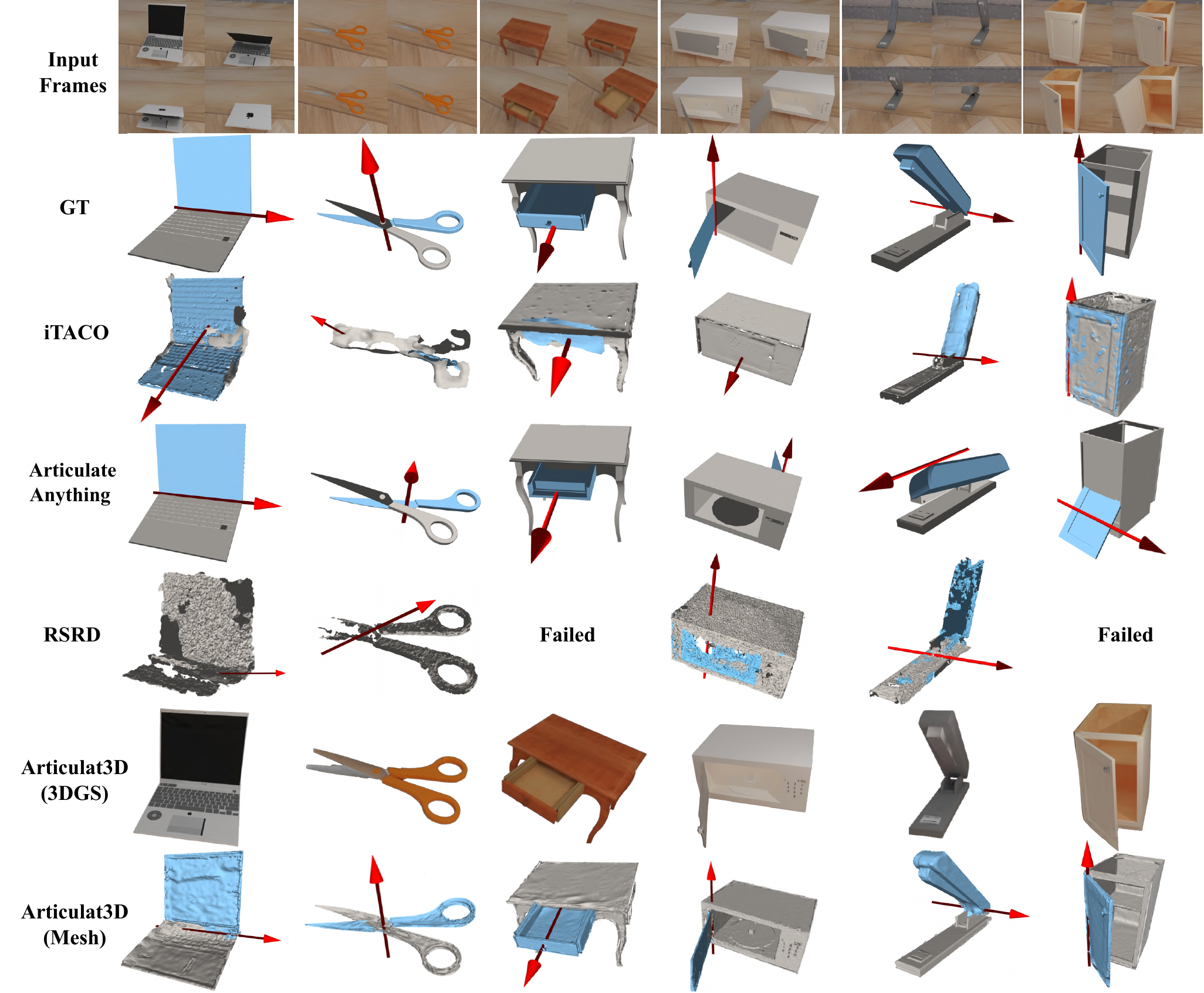}
    \vspace{-10pt}
    \caption{\textbf{Visual results} on Video2Articulation-S dataset~\cite{peng2025itaco}.}
    \label{fig:video2art}
\end{figure}

\begin{figure}[t]
    \centering
    \includegraphics[width=1.0\linewidth]{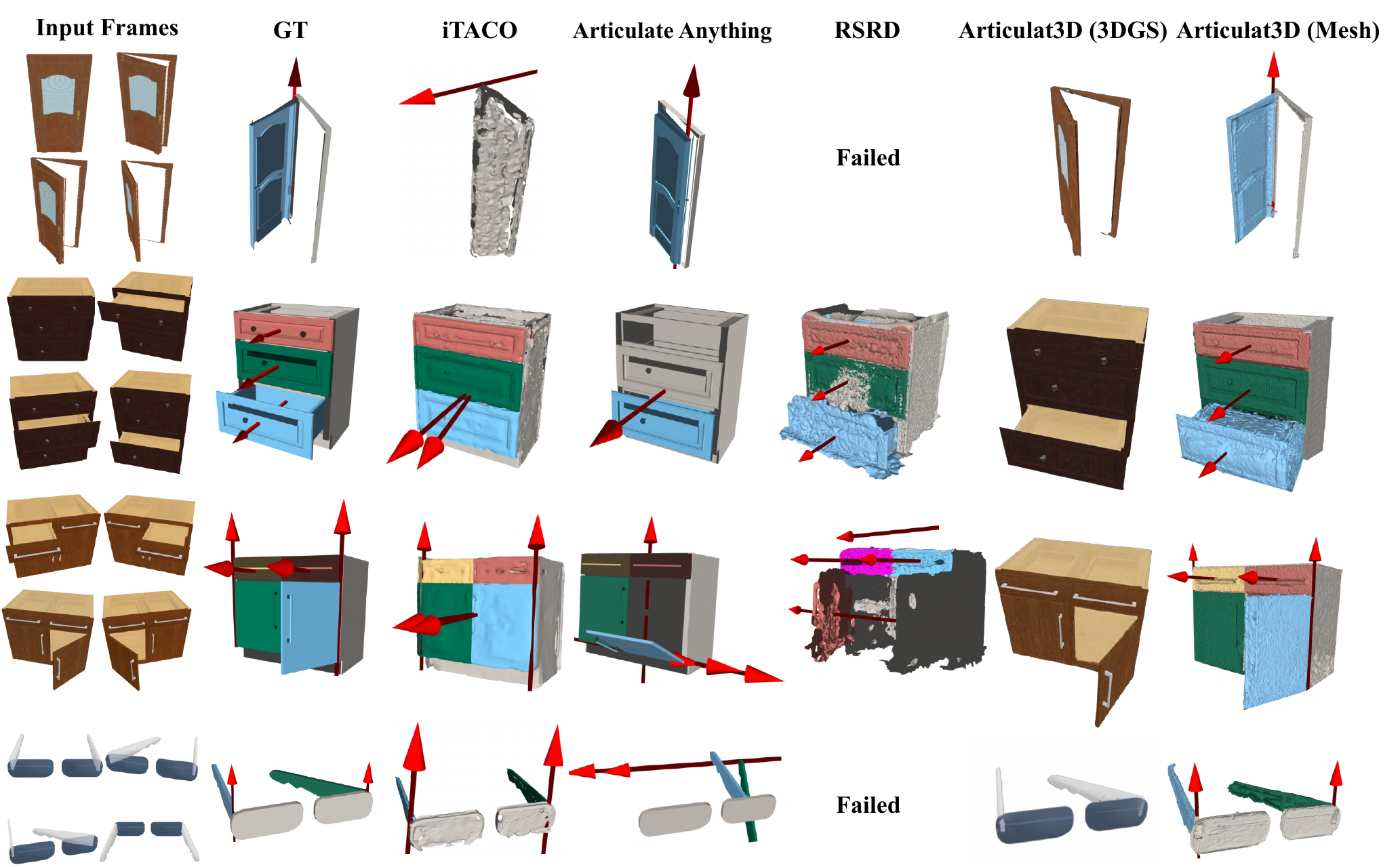}
    \vspace{-10pt}
    \caption{\textbf{Visual results} on Articulat3D-Sim dataset.}
    \label{fig:articulat3d-sim}
\end{figure}

\section{Experiments}
\subsection{Datasets}
We conduct a comprehensive evaluation on three distinct datasets to assess the performance of existing methods on objects with varying articulation complexity.

\noindent
\textbf{Video2Articulation-S.} It is a dataset proposed by~\citep{peng2025itaco}, which serves as our benchmark for simple articulated objects. It consists of 73 test videos across 11 categories of synthetic objects, where each object has only a single movable part.

\noindent
\textbf{Articulat3D-Sim.} 
We introduce this dataset of 20 videos across 17 categories from PartNet-Mobility dataset~\cite{xiang2020sapien} to evaluate complex multi-part kinematics (ranging from 2 to 7 parts). Each sequence features natural interaction, combining synchronized articulations with significant camera motion.

\noindent
\textbf{Articulat3D-Real.} 
To evaluate performance in real world, we introduce this dataset, which consists of 12 sequences of 9 distinct categories (\eg, cabinets, and doors). Captured using an iPhone 16 Pro Max, these sequences feature diverse lighting conditions and complex background clutter, providing a benchmark for assessing the generalizability of articulated reconstruction in the wild.

\subsection{Evaluation Metrics}
We evaluate \our across four dimensions: 
(1) \textbf{Joint Estimation}: Axis Err ($^\circ$) and Position Err (cm) for kinematic accuracy;
(2) \textbf{Reconstruction}: Chamfer Distance (cm) for the whole object (CD-w), movable parts (CD-m), and static base (CD-s);
(3) \textbf{Tracking}: End-Point Error (EPE) for temporal motion coherence;
(4) \textbf{View Synthesis}: PSNR, SSIM, and LPIPS for photometric quality.
For qualitative results, we visualize reconstruction results as meshes, using grey for static components, various colors for dynamic parts, and red arrows for joint axes.
Detailed definitions are provided in our Supp. Mat.

\subsection{Results}
We compare \our against four state-of-the-art baselines: (1) \textbf{RSRD}~\cite{kerr2024robot}, which optimizes part-level $SE(3)$ transformations; (2) \textbf{Articulate Anything}~\cite{le2024articulate}, a foundation model that retrieves joint parameters from a mesh library; (3) \textbf{iTACO}~\cite{peng2025itaco}, which estimates articulation via point cloud motion analysis; (4) \textbf{Shape of Motion}~\cite{wang2025shape}, a 4D-GS based method representing dynamics with generic motion bases. For a fair comparison, RSRD~\cite{kerr2024robot} is adapted with NKSR mesh reconstruction, Articulate Anything~\cite{le2024articulate} is provided with the PartNet-Mobility library, and iTACO~\cite{peng2025itaco} is given ground-truth depth. We provide exhaustive implementation details and configuration settings for all baselines in Supp. Mat.

For quantitative results, N/A entries in Tab.~\ref{tab:exp} indicate inapplicable metrics: EPE is omitted for datasets lacking ground-truth trajectories; view synthesis metrics are excluded for Articulate Anything~\cite{le2024articulate} and iTACO~\cite{peng2025itaco} as they do not support rendering. For qualitative results, we visualize reconstruction results as meshes, using grey for static components, various colors for dynamic parts, and red arrows for joint axes. Failed indicates cases where RSRD~\cite{kerr2024robot} failed to reconstruct geometry. Since baseline methods such as Articulate Anything~\cite{le2024articulate} and iTACO~\cite{peng2025itaco} only provide static meshes rather than per-frame poses, their results are shown in fixed configurations that may not match the GT states.

\noindent
\textbf{Quantitative results.} 
As shown in Tab.~\ref{tab:exp}, \our consistently outperforms all baselines. 
Regarding \textbf{Joint Estimation}, \our achieves best precision with an average axis error of $1.60^\circ$ on Video2Articulation-S and $0.53^\circ$ on Articulat3D-Sim. In contrast, iTACO~\cite{peng2025itaco} fails to handle multi-part objects ($48.50^\circ$ axis error), highlighting its limitations in complex kinematic scenarios. Regarding \textbf{Reconstruction Fidelity}, Articulate Anything~\cite{le2024articulate} and RSRD~\cite{kerr2024robot} suffer from catastrophic position errors ($>90$ cm) due to template mismatch or structural collapse. Conversely, \our still maintains sub-centimeter accuracy and high-fidelity synthesis. While Shape of Motion~\cite{wang2025shape} is a dedicated dynamic renderer, its motion bases lead to significant geometric drift. In contrast, our kinematic refinement enforces strict rigid-body laws, ensuring both high-fidelity synthesis ($37.80$ PSNR) and physically plausible 3D trajectories. 



\begin{figure}[t]
    \centering
    \includegraphics[width=1.0\linewidth]{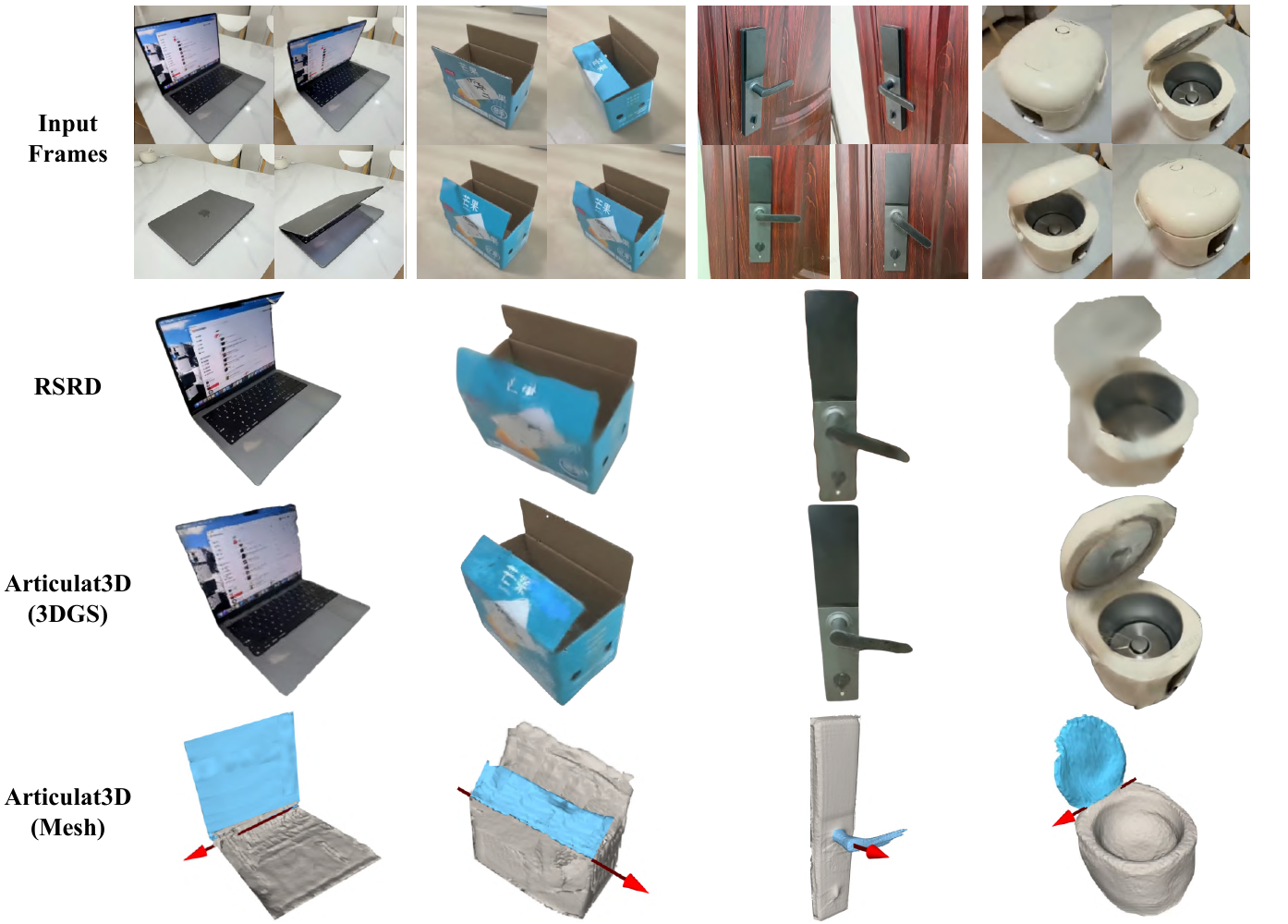}
      \vspace{-5pt}
    \caption{\textbf{Visual results} on Articulat3D-Real dataset.}
    \label{fig:articulat3d-real}
\end{figure}

\noindent
\textbf{Qualitative Results.}
Visual comparisons in Fig.~\ref{fig:video2art},~\ref{fig:articulat3d-sim}, and~\ref{fig:articulat3d-real} confirm our quantitative gains. 
In \textbf{Video2Articulation-S} (Fig.~\ref{fig:video2art}), \our preserves structural rigidity by filtering tracking noise. 
On more challenging \textbf{Articulat3D-Sim} dataset (Fig.~\ref{fig:articulat3d-sim}), our framework effectively reconstructs complex multi-part kinematics. While concurrent methods often suffer from ghosting or part-intersections, our two-stage optimization ensures clean, axis-aligned trajectories. Finally, results on real world dataset, \textbf{Articulat3D-Real} (Fig.~\ref{fig:articulat3d-real}) demonstrate our robustness to in-the-wild challenges, including handheld jitter, cluttered backgrounds, and significant hand occlusions. \our still produces high-fidelity digital twins that are inherently ready for physical interaction.


\begin{table}[t]
 \centering
    \caption{\textbf{Ablation Study} on Articulat3D-Sim dataset.}
    \vspace{-7pt}
    \label{tab:ablation}
    \scriptsize
    \setlength{\tabcolsep}{1.5pt}
    \resizebox{\linewidth}{!}{
        \begin{tabular}{cccccccccc}
        \toprule
        \multirow{2}{*}{Method} & \multicolumn{2}{c}{Joint Estimation} & \multicolumn{3}{c}{Reconstruction} & \multicolumn{1}{c}{Tracking} & \multicolumn{3}{c}{View Synthesis} \\
        \cmidrule(lr){2-3} \cmidrule(lr){4-6} \cmidrule(lr){7-7} \cmidrule(lr){8-10}
        & Axis Err $\downarrow$ & Pos Err $\downarrow$ & CD-w $\downarrow$ & CD-m $\downarrow$ & CD-s $\downarrow$ & EPE $\downarrow$ & PSNR $\uparrow$ & SSIM $\uparrow$ & LPIPS $\downarrow$ \\
        \midrule
        \rowcolor{gray!10}\our w/o motion prior & 43.42 & 31.26 & 18.78 & 92.45 & 17.73 & 28.29 & 20.78 & 0.95 & 0.07 \\
        \our w/noise & 0.56 & 0.67 & 0.78 & 0.88 & 0.86 & 0.05 & 37.62 & 0.98 & 0.03 \\
        \rowcolor{gray!10}\our w/o prior init & 1.36 & 3.42 & 1.13 & 1.67 & 0.96 & 0.06 & 30.89 & 0.98 & 0.04\\
        \our w/o kinem. const. & 0.86 & 4.98 & 5.19 & 7.71 & 2.76 & 0.09 & 35.45 & 0.97 & 0.02     \\
        \rowcolor{gray!10}\our w less data & 0.61 & 0.68 & 0.83 & 0.92 & 0.86 & 0.07 & 34.60 & 0.97 & 0.03\\
        \rowcolor[HTML]{D7F6FF} \textbf{\our} & \textbf{0.53} & \textbf{0.65} & \textbf{0.76} & \textbf{0.84} & \textbf{0.85} & \textbf{0.04} & \textbf{37.80} & \textbf{0.98} & \textbf{0.03} \\
        \bottomrule
        \end{tabular}
    }
    \vspace{-5pt}
\end{table}

\begin{figure}[t]
  \centering
  \begin{minipage}[t]{0.45\textwidth}
    \centering
    \includegraphics[width=0.75\linewidth]{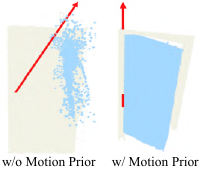} %
    \vspace{-7pt}
    \caption{\textbf{Visual impact of motion prior.} Without 3D motion guidance, motion-geometry ambiguity leads to distorted structures.}
    \label{fig:no_track}
  \end{minipage}
  \hfill 
  \begin{minipage}[t]{0.50\textwidth}
    \centering
    \includegraphics[width=\linewidth]{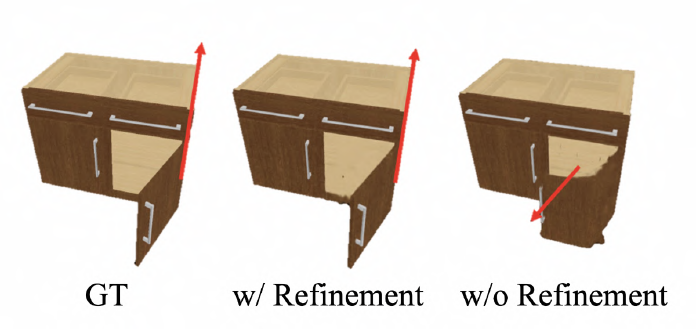} 
    \vspace{-17pt}
    \caption{\textbf{Visual impact of refinement.} Without our initialization, the model converges to local minima with misaligned joint axes and distorted geometry.}
    \label{fig:ablation_refine}
  \end{minipage}
\end{figure}

\subsection{Ablation Study}
\noindent 
\textbf{Motion Prior Guidance.}
Removing 3D track guidance leads to performance drop (\our w/o motion prior) in Tab.~\ref{tab:ablation} and visible artifacts, as shown in Fig.~\ref{fig:no_track}. This prior is essential for resolving the inherent ambiguity between geometry and kinematics and decoupling local geometry from global motion dynamics. Our full model successfully resolves this coupling, yielding stable and accurate digital twins.

\begin{figure}[t]
    \centering
    \includegraphics[width=0.85\linewidth]{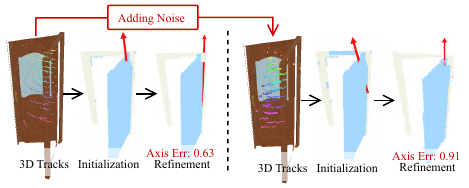}
    \vspace{-5pt}
    \caption{\textbf{Robustness to noisy motion priors.} Our kinematic refinement stage recovers accurate part segmentation and joint parameters even when initialized with highly inconsistent 3D tracks.}
    \label{fig:noisy_traj}
\end{figure}

\noindent 
\textbf{Robustness to noise in Motion Priors.}
By injecting Gaussian noise into raw 3D trajectories (\our w/ noise), we demonstrate that \our remains robust, as shown in Tab.~\ref{tab:ablation} and Fig.~\ref{fig:noisy_traj}. Our explicit axis-angle parameterization acts as a rigid kinematic bottleneck that filters non-physical noise. Instead of directly supervising on inconsistent raw coordinates, this formulation forces motion onto a valid kinematic manifold, effectively refining and correcting the upstream trajectories in a geometrically consistent manner.

\noindent 
\textbf{Low-Dimensional Motion Bases Initialization.}
We evaluate the necessity of Motion Prior–Driven Initialization by attempting to initialize kinematic parameters directly from raw 3D tracks (\our w/o prior init). While explicit kinematic primitives provide strong physical grounding, their optimization is inherently non-convex and highly sensitive to initial values. As shown in Tab.~\ref{tab:ablation} and Fig.~\ref{fig:ablation_refine}, bypassing this stage causes the model to converge to poor local minima, characterized by misaligned part segmentation and inaccurate joint axes. This initialization stage exploits the low-dimensional structure of articulated motion to transform an ill-posed monocular tracking problem into a well-posed kinematic refinement task, ensuring temporal coherence and geometric accuracy.

\begin{figure}
    \centering
    \includegraphics[width=0.90\linewidth]{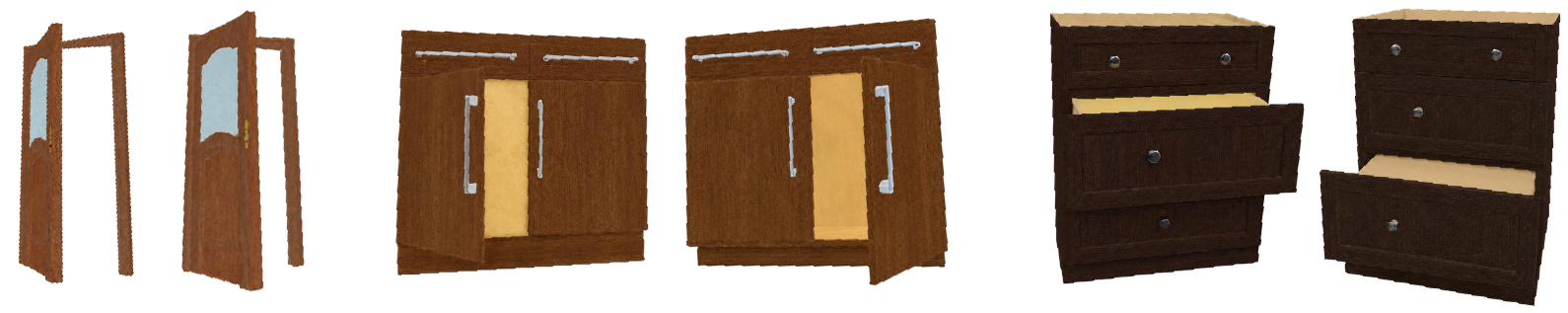}
    \vspace{-7pt}
    \caption{\textbf{Novel-view synthesis results on the Articulat3D-Sim dataset.} The displayed viewpoints are not observed during training.}
    \vspace{-10pt}
    \label{fig:supp_novel_view}
\end{figure}

\noindent 
\textbf{Novel-view synthesis on synthetic data.}
We additionally render the reconstructed articulated objects from viewpoints that are not observed during
training. As shown in Fig~\ref{fig:supp_novel_view}, Articulat3D produces visually consistent novel-view renderings while preserving object appearance, geometric structure, and articulation details. This indicates that the learned representation captures a coherent 3D articulated object rather than merely reproducing the observed training views from the input sequence directly.

\noindent
\textbf{Geometric and Motion Constraints Refinement.}
Removing the geometric and motion refinement stage leads to a marked degradation in joint accuracy and structural fidelity (\our w/o kinem. const.), as shown in Tab.~\ref{tab:ablation}. Without the explicit kinematic bottleneck, the model fails to enforce rigid-body physical laws, resulting in "floating" components and geometric drift, as shown in Fig.~\ref{fig:ablation_refine}. This is evidenced by the sharp increase in Position Error (from 0.65 to 4.98) and CD-m (from 0.84 to 7.71), which confirm that our refinement stage is indispensable for transforming initial motion estimates into physically plausible and temporally coherent digital twins.

\begin{figure}[t]
    \centering
    \includegraphics[width=0.85\linewidth]{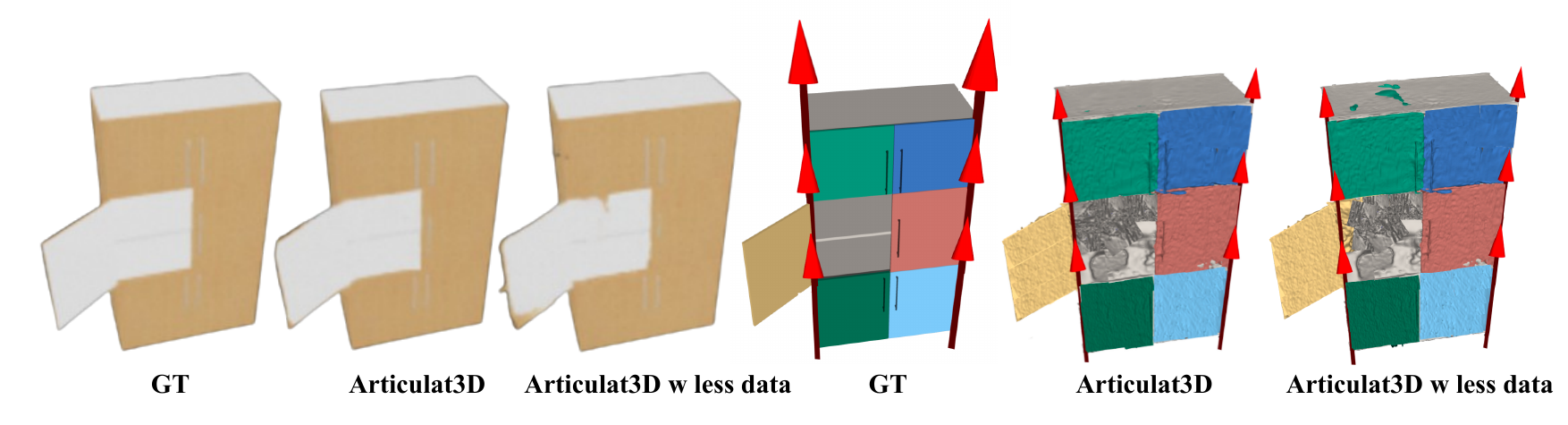}
      \vspace{-7pt}
    \caption{\textbf{Robustness to data sparsity.} Comparison the results between GT, \our trained with full frames, and \our trained with 1/3 sub-sampled frames.}
    \label{fig:less-data}
\end{figure}

\noindent
\textbf{Robustness to data sparsity.} 
To evaluate the efficiency and robustness of our under limited visual observations, we conduct an experiment by uniformly subsampling the input video to use only 1/3 of the original frames. As reported in Tab.~\ref{tab:ablation} and Fig.~\ref{fig:less-data}, while the rendering quality experiences a moderate decline (PSNR decreases from 37.80 to 34.60) due to fewer photometric constraints for Gaussian optimization, the kinematic accuracy remains remarkably stable. Specifically, the Axis Error and Position Error only show marginal increases ($0.53^\circ \to 0.61^\circ$ and $0.65 \to 0.68$\,cm, respectively). This demonstrates that our Motion Prior-Driven Initialization effectively captures the low-dimensional structure of articulation even from temporally sparse data, and the Geometric Constraints Refinement successfully maintains structural integrity despite the reduced supervision.

\section{Conclusion}
We present \textbf{\our}, a framework for reconstructing physically plausible, interactable digital twins of articulated objects from a single casually captured monocular video, even when motion begins immediately. A motion-prior-driven initialization combined with rigorous geometric constraint refinement successfully regularizes the ill-posed monocular reconstruction problem into a structured optimization of kinematic primitives. Without requiring static scans or controlled environments, \our offers a scalable, robust solution for modeling the articulated world and enabling advanced robotic manipulation applications.

\noindent
\textbf{Limitation and future work. } 
\our’s performance may degrade on texture-less surfaces where sparse 3D tracks become unreliable. Moreover, our current implementation assumes independent joints rather than hierarchical kinematic structures, which limits the modeling of nested assemblies. Future work will address these issues by integrating multi-view geometric priors and exploring the estimation of dynamic physical parameters (\eg, mass and friction).

\section*{Acknowledgements}

This work was partly supported by the Industry-University-Research Innovation Fund of the Ministry of Education (Grant No. 2025XJS023), the Central Guidance for Local Science and Technology Development Fund(Grant No. ZYYD2025QY19).

%
%
\bibliographystyle{splncs04}
\bibliography{root}

@string{cvpr = "Proc. of IEEE Conf. on Computer Vision and Pattern Recognition"}

@string{iccv = "Proc. of IEEE Intl. Conf. on Computer Vision"}

@string{nips = "Proc. of Advances in Neural Information Processing Systems"}

@string{iclr = "Proc. of International Conference on Learning Representations"}

@string{threedv = "Proc. of Intl. Conf. on 3D Vision"}

@inproceedings{weng2024neural,
  title={Neural implicit representation for building digital twins of unknown articulated objects},
  author={Weng, Yijia and Wen, Bowen and Tremblay, Jonathan and Blukis, Valts and Fox, Dieter and Guibas, Leonidas and Birchfield, Stan},
  booktitle=cvpr,
  pages={3141--3150},
  year={2024}
}

@inproceedings{luo2025physpart,
  title={Physpart: Physically plausible part completion for interactable objects},
  author={Luo, Rundong and Geng, Haoran and Deng, Congyue and Li, Puhao and Wang, Zan and Jia, Baoxiong and Guibas, Leonidas and Huang, Siyuan},
  booktitle={IEEE International Conference on Robotics and Automation},
  pages={12386--12393},
  year={2025},
  organization={IEEE}
}

@inproceedings{liu2024cage,
  title={Cage: Controllable articulation generation},
  author={Liu, Jiayi and Tam, Hou In Ivan and Mahdavi-Amiri, Ali and Savva, Manolis},
  booktitle=cvpr,
  pages={17880--17889},
  year={2024}
}

@inproceedings{zhao2025physsplat,
  title={PhysSplat: Efficient Physics Simulation for 3D Scenes via MLLM-Guided Gaussian Splatting},
  author={Zhao, Haoyu and Wang, Hao and Zhao, Xingyue and Fei, Hao and Wang, Hongqiu and Long, Chengjiang and Zou, Hua},
  booktitle=iccv,
  pages={5242--5252},
  year={2025}
}

@article{zhao2025high,
  title={High-fidelity simulated data generation for real-world zero-shot robotic manipulation learning with gaussian splatting},
  author={Zhao, Haoyu and Zeng, Cheng and Zhuang, Linghao and Zhao, Yaxi and Xue, Shengke and Wang, Hao and Zhao, Xingyue and Li, Zhongyu and Li, Kehan and Huang, Siteng and others},
  journal={arXiv preprint arXiv:2510.10637},
  year={2025}
}

@inproceedings{yang2023reconstructing,
  title={Reconstructing animatable categories from videos},
  author={Yang, Gengshan and Wang, Chaoyang and Reddy, N Dinesh and Ramanan, Deva},
  booktitle=cvpr,
  pages={16995--17005},
  year={2023}
}

@article{liu2025videoartgs,
  title={VideoArtGS: Building Digital Twins of Articulated Objects from Monocular Video},
  author={Liu, Yu and Jia, Baoxiong and Lu, Ruijie and Gan, Chuyue and Chen, Huayu and Ni, Junfeng and Zhu, Song-Chun and Huang, Siyuan},
  journal={arXiv preprint arXiv:2509.17647},
  year={2025}
}

@article{torne2024reconciling,
  title={Reconciling reality through simulation: A real-to-sim-to-real approach for robust manipulation},
  author={Torne, Marcel and Simeonov, Anthony and Li, Zechu and Chan, April and Chen, Tao and Gupta, Abhishek and Agrawal, Pulkit},
  journal={arXiv preprint arXiv:2403.03949},
  year={2024}
}

@article{kerr2024robot,
  title={Robot see robot do: Imitating articulated object manipulation with monocular 4d reconstruction},
  author={Kerr, Justin and Kim, Chung Min and Wu, Mingxuan and Yi, Brent and Wang, Qianqian and Goldberg, Ken and Kanazawa, Angjoo},
  journal={arXiv preprint arXiv:2409.18121},
  year={2024}
}

@article{mandi2024real2code,
  title={{Real2Code}: Reconstruct articulated objects via code generation},
  author={Mandi, Zhao and Weng, Yijia and Bauer, Dominik and Song, Shuran},
  journal={arXiv preprint arXiv:2406.08474},
  year={2024}
}

@article{le2024articulate,
  title={Articulate-anything: Automatic modeling of articulated objects via a vision-language foundation model},
  author={Le, Long and Xie, Jason and Liang, William and Wang, Hung-Ju and Yang, Yue and Ma, Yecheng Jason and Vedder, Kyle and Krishna, Arjun and Jayaraman, Dinesh and Eaton, Eric},
  journal=iclr,
  year={2024}
}

@inproceedings{jiang2022ditto,
  title={Ditto: Building digital twins of articulated objects from interaction},
  author={Jiang, Zhenyu and Hsu, Cheng-Chun and Zhu, Yuke},
  booktitle=cvpr,
  pages={5616--5626},
  year={2022}
}

@inproceedings{liu2025building,
  title={Building interactable replicas of complex articulated objects via gaussian splatting},
  author={Liu, Yu and Jia, Baoxiong and Lu, Ruijie and Ni, Junfeng and Zhu, Song-Chun and Huang, Siyuan},
  booktitle=iclr,
  year={2025}
}

@inproceedings{liu2023paris,
  title={Paris: Part-level reconstruction and motion analysis for articulated objects},
  author={Liu, Jiayi and Mahdavi-Amiri, Ali and Savva, Manolis},
  booktitle=iccv,
  pages={352--363},
  year={2023}
}

@inproceedings{peng2025itaco,
    author = {Weikun Peng and Jun Lv and Cewu Lu and Manolis Savva},
    title = {{iTACO: Interactable Digital Twins of Articulated Objects from Casually Captured RGBD Videos}},
    booktitle = threedv,
    year = {2025}
}

@article{zhang2025tapip3d,
  title={{TAPIP3D}: Tracking Any Point in Persistent 3D Geometry},
  author={Zhang, Bowei and Ke, Lei and Harley, Adam W and Fragkiadaki, Katerina},
  journal={arXiv preprint arXiv:2504.14717},
  year={2025}
}

@inproceedings{wang2025shape,
  title={Shape of motion: 4d reconstruction from a single video},
  author={Wang, Qianqian and Ye, Vickie and Gao, Hang and Zeng, Weijia and Austin, Jake and Li, Zhengqi and Kanazawa, Angjoo},
  booktitle=cvpr,
  pages={9660--9672},
  year={2025}
}

@inproceedings{guo2025articulatedgs,
  title={{ArticulatedGS}: Self-supervised digital twin modeling of articulated objects using 3d gaussian splatting},
  author={Guo, Junfu and Xin, Yu and Liu, Gaoyi and Xu, Kai and Liu, Ligang and Hu, Ruizhen},
  booktitle={Proceedings of the Computer Vision and Pattern Recognition Conference},
  pages={27144--27153},
  year={2025}
}

@article{teed2021droid,
  title={Droid-slam: Deep visual slam for monocular, stereo, and rgb-d cameras},
  author={Teed, Zachary and Deng, Jia},
  journal=nips,
  volume={34},
  pages={16558--16569},
  year={2021}
}

@inproceedings{li2025megasam,
  title={{MegaSaM}: Accurate, fast and robust structure and motion from casual dynamic videos},
  author={Li, Zhengqi and Tucker, Richard and Cole, Forrester and Wang, Qianqian and Jin, Linyi and Ye, Vickie and Kanazawa, Angjoo and Holynski, Aleksander and Snavely, Noah},
  booktitle=cvpr,
  pages={10486--10496},
  year={2025}
}

@inproceedings{wang2024dust3r,
  title={Dust3r: Geometric 3d vision made easy},
  author={Wang, Shuzhe and Leroy, Vincent and Cabon, Yohann and Chidlovskii, Boris and Revaud, Jerome},
  booktitle=cvpr,
  pages={20697--20709},
  year={2024}
}

@article{zhang2024monst3r,
  title={Monst3r: A simple approach for estimating geometry in the presence of motion},
  author={Zhang, Junyi and Herrmann, Charles and Hur, Junhwa and Jampani, Varun and Darrell, Trevor and Cole, Forrester and Sun, Deqing and Yang, Ming-Hsuan},
  journal=iclr,
  year={2025}
}

@inproceedings{wang2025continuous,
  title={Continuous 3d perception model with persistent state},
  author={Wang, Qianqian and Zhang, Yifei and Holynski, Aleksander and Efros, Alexei A and Kanazawa, Angjoo},
  booktitle=cvpr,
  pages={10510--10522},
  year={2025}
}

@inproceedings{duan20244d,
  title={4d-rotor gaussian splatting: towards efficient novel view synthesis for dynamic scenes},
  author={Duan, Yuanxing and Wei, Fangyin and Dai, Qiyu and He, Yuhang and Chen, Wenzheng and Chen, Baoquan},
  booktitle={ACM SIGGRAPH 2024 Conference Papers},
  pages={1--11},
  year={2024}
}

@inproceedings{luiten2024dynamic,
  title={Dynamic 3d gaussians: Tracking by persistent dynamic view synthesis},
  author={Luiten, Jonathon and Kopanas, Georgios and Leibe, Bastian and Ramanan, Deva},
  booktitle=threedv,
  pages={800--809},
  year={2024},
}

@inproceedings{wu20244d,
  title={4d gaussian splatting for real-time dynamic scene rendering},
  author={Wu, Guanjun and Yi, Taoran and Fang, Jiemin and Xie, Lingxi and Zhang, Xiaopeng and Wei, Wei and Liu, Wenyu and Tian, Qi and Wang, Xinggang},
  booktitle=cvpr,
  pages={20310--20320},
  year={2024}
}

@article{yang2023real,
  title={Real-time photorealistic dynamic scene representation and rendering with 4d gaussian splatting},
  author={Yang, Zeyu and Yang, Hongye and Pan, Zijie and Zhang, Li},
  journal=iclr,
  year={2024}
}

@inproceedings{heppert2023carto,
  title={Carto: Category and joint agnostic reconstruction of articulated objects},
  author={Heppert, Nick and Irshad, Muhammad Zubair and Zakharov, Sergey and Liu, Katherine and Ambrus, Rares Andrei and Bohg, Jeannette and Valada, Abhinav and Kollar, Thomas},
  booktitle=cvpr,
  pages={21201--21210},
  year={2023}
}

@inproceedings{wei2022self,
  title={Self-supervised neural articulated shape and appearance models},
  author={Wei, Fangyin and Chabra, Rohan and Ma, Lingni and Lassner, Christoph and Zollh{\"o}fer, Michael and Rusinkiewicz, Szymon and Sweeney, Chris and Newcombe, Richard and Slavcheva, Mira},
  booktitle={Proceedings of the IEEE/CVF Conference on Computer Vision and Pattern Recognition},
  pages={15816--15826},
  year={2022}
}

@article{goyal2025geopard,
  title={{GEOPARD}: Geometric Pretraining for Articulation Prediction in 3D Shapes},
  author={Goyal, Pradyumn and Petrov, Dmitry and Andrews, Sheldon and Ben-Shabat, Yizhak and Liu, Hsueh-Ti Derek and Kalogerakis, Evangelos},
  journal={arXiv preprint arXiv:2504.02747},
  year={2025}
}

@inproceedings{xia2025drawer,
  title={Drawer: Digital reconstruction and articulation with environment realism},
  author={Xia, Hongchi and Su, Entong and Memmel, Marius and Jain, Arhan and Yu, Raymond and Mbiziwo-Tiapo, Numfor and Farhadi, Ali and Gupta, Abhishek and Wang, Shenlong and Ma, Wei-Chiu},
  booktitle={Proceedings of the Computer Vision and Pattern Recognition Conference},
  pages={21771--21782},
  year={2025}
}

@article{hu2017learning,
  title={Learning to predict part mobility from a single static snapshot},
  author={Hu, Ruizhen and Li, Wenchao and Van Kaick, Oliver and Shamir, Ariel and Zhang, Hao and Huang, Hui},
  journal={ACM Transactions On Graphics (TOG)},
  volume={36},
  number={6},
  pages={1--13},
  year={2017},
  publisher={ACM New York, NY, USA}
}

@article{yi2018deep,
  title={Deep part induction from articulated object pairs},
  author={Yi, Li and Huang, Haibin and Liu, Difan and Kalogerakis, Evangelos and Su, Hao and Guibas, Leonidas},
  journal={arXiv preprint arXiv:1809.07417},
  year={2018}
}

@article{wu2025reartgs,
  title={{REArtGS}: Reconstructing and generating articulated objects via 3d gaussian splatting with geometric and motion constraints},
  author={Wu, Di and Liu, Liu and Linli, Zhou and Huang, Anran and Song, Liangtu and Yu, Qiaojun and Wu, Qi and Lu, Cewu},
  journal={arXiv preprint arXiv:2503.06677},
  year={2025}
}

@article{wu2025reartgs++,
  title={{REArtGS++}: Generalizable Articulation Reconstruction with Temporal Geometry Constraint via Planar Gaussian Splatting},
  author={Wu, Di and Liu, Liu and Huang, Anran and Liu, Yuyan and Yu, Qiaojun and Liu, Shaofan and Song, Liangtu and Lu, Cewu},
  journal={arXiv preprint arXiv:2511.17059},
  year={2025}
}

@inproceedings{song2024reacto,
  title={Reacto: Reconstructing articulated objects from a single video},
  author={Song, Chaoyue and Wei, Jiacheng and Foo, Chuan Sheng and Lin, Guosheng and Liu, Fayao},
  booktitle=cvpr,
  pages={5384--5395},
  year={2024}
}

@inproceedings{zhou2025monomobility,
  title={Monomobility: Zero-shot 3d mobility analysis from monocular videos},
  author={Zhou, Hongyi and Guo, Yulan and Wang, Xiaogang and Xu, Kai},
  booktitle=cvpr,
  pages={8800--8809},
  year={2025}
}

@article{carion2025sam,
  title={Sam 3: Segment anything with concepts},
  author={Carion, Nicolas and Gustafson, Laura and Hu, Yuan-Ting and Debnath, Shoubhik and Hu, Ronghang and Suris, Didac and Ryali, Chaitanya and Alwala, Kalyan Vasudev and Khedr, Haitham and Huang, Andrew and others},
  journal={arXiv preprint arXiv:2511.16719},
  year={2025}
}

@article{wu2025qwen,
  title={Qwen-image technical report},
  author={Wu, Chenfei and Li, Jiahao and Zhou, Jingren and Lin, Junyang and Gao, Kaiyuan and Yan, Kun and Yin, Sheng-ming and Bai, Shuai and Xu, Xiao and Chen, Yilei and others},
  journal={arXiv preprint arXiv:2508.02324},
  year={2025}
}

@inproceedings{xiang2020sapien,
  title={Sapien: A simulated part-based interactive environment},
  author={Xiang, Fanbo and Qin, Yuzhe and Mo, Kaichun and Xia, Yikuan and Zhu, Hao and Liu, Fangchen and Liu, Minghua and Jiang, Hanxiao and Yuan, Yifu and Wang, He and others},
  booktitle=cvpr,
  pages={11097--11107},
  year={2020}
}

@article{zhao2024hfgs,
  title={Hfgs: 4d gaussian splatting with emphasis on spatial and temporal high-frequency components for endoscopic scene reconstruction},
  author={Zhao, Haoyu and Zhao, Xingyue and Zhu, Lingting and Zheng, Weixi and Xu, Yongchao},
  journal={arXiv preprint arXiv:2405.17872},
  year={2024}
}

@article{zhao2024sg,
  title={Sg-gs: Photo-realistic animatable human avatars with semantically-guided gaussian splatting},
  author={Zhao, Haoyu and Yang, Chen and Wang, Hao and Zhao, Xingyue and Shen, Wei},
  journal={arXiv preprint arXiv:2408.09665},
  volume={2},
  number={3},
  year={2024}
}

@article{zhao2025towards,
  title={Towards affordance-aware robotic dexterous grasping with human-like priors},
  author={Zhao, Haoyu and Zhuang, Linghao and Zhao, Xingyue and Zeng, Cheng and Xu, Haoran and Jiang, Yuming and Cen, Jun and Wang, Kexiang and Guo, Jiayan and Huang, Siteng and others},
  journal={arXiv preprint arXiv:2508.08896},
  year={2025}
}

@article{zhao20243d,
  title={3D-Consistent Human Avatars with Sparse Inputs via Gaussian Splatting and Contrastive Learning},
  author={Zhao, Haoyu and Wang, Hao and Yang, Chen and Shen, Wei},
  journal={arXiv preprint arXiv:2408.09663},
  year={2024}
}

\newpage
\appendix
\onecolumn
\section{Metrics}
\label{sec:supp_metrics}
This section provides the formal definitions for the metrics presented in Tab.~\ref{tab:exp}:

\begin{itemize}
    \item \textbf{Joint Estimation:} We evaluate kinematics using \textbf{Axis Err}, the angular deviation ($^\circ$) between predicted and ground-truth joint axes, and \textbf{Position Err}, the Euclidean distance (cm) between predicted and ground-truth pivot points.
    
    \item \textbf{Reconstruction:} We compute the bi-directional Chamfer Distance (CD) on 10,000 points sampled from the reconstructed and ground-truth meshes. We report CD (cm) for the entire object (\textbf{CD-w}), movable components (\textbf{CD-m}), and the static base (\textbf{CD-s}) to assess structural fidelity.
    
    \item \textbf{3D Tracking:} Following~\cite{wang2025shape}, \textbf{EPE} (End-Point Error) measures the average 3D Euclidean distance between predicted and ground-truth trajectories across all frames, quantifying the model's ability to capture continuous motion.
    
    \item \textbf{View Synthesis:} We use \textbf{PSNR}, \textbf{SSIM}, and \textbf{LPIPS} to evaluate novel view rendering. These metrics ensure that our kinematic constraints maintain the high-fidelity rendering performance of the underlying Gaussian Splatting framework.
\end{itemize}

\section{Implementation and Training Details}
\subsection{Motion Prior Grouping via K-means Clustering}
\label{sec:supp_kmeans}
To provide a robust initialization for the motion coefficients $\mathbf{w}_i$ in Sec.~\ref{sec:geo_motion_refine}, we propose a spatio-temporal trajectory clustering strategy. Unlike standard spatial K-means, which ignores the temporal evolutionary structure of articulated motion, our approach exploits both the temporal motion directions and the canonical spatial layout of 3D trajectories. Importantly, this initialization does not assume that different articulated parts must move in disjoint temporal windows.

\noindent
\textbf{Kinematic Priors.}
Our grouping algorithm is guided by three priors derived from common articulated interactions:
\begin{itemize}
    \item \textbf{Motion Direction Consistency:} Gaussians belonging to the same rigid part tend to exhibit similar velocity-direction patterns over time.
    \item \textbf{Spatial Coherence:} Gaussians from the same physical part are expected to be spatially coherent in the canonical frame.
    \item \textbf{Static-Dynamic Separation:} Nearly static Gaussians, if identified by the tracking prior, are assigned to the static base, while the remaining dynamic Gaussians are clustered into movable part candidates.
\end{itemize}

\noindent
\textbf{Weighted Motion Energy Calculation.}
Given a set of 3D trajectories $\mathcal{X} = \{\mathbf{x}_{i,t}\}$ with associated visibility $\mathbf{v}_{i,t}$ and confidence scores $\mathbf{c}_{i,t}$, we first perform linear interpolation to handle occlusions. We then define the weighted speed $\tilde{s}_{i,t}$ for each Gaussian $i$ at interval $t$ as:
\begin{equation}
    \tilde{s}_{i,t} = \|\mathbf{x}_{i,t+1} - \mathbf{x}_{i,t}\|_2, 
    \quad w_{i,t} = (\mathbf{v}_{i,t} \cdot \mathbf{v}_{i,t+1}) \cdot \frac{\mathbf{c}_{i,t} + \mathbf{c}_{i,t+1}}{2}
\end{equation}

To characterize the motion of each Gaussian, we use both its velocity direction and its spatial position as clustering cues. Specifically, for each Gaussian $i$ at interval $t$, we compute the normalized velocity direction as:
\begin{equation}
    \mathbf{s}_{i,t}=w_{i,t}  \cdot \frac{\mathbf{x}_{i,t+1}-\mathbf{x}_{i,t}}{\tilde{s}_{i,t}+\epsilon}
\end{equation}
Meanwhile, we use the spatial position of each Gaussian in the canonical frame as its geometric cue:
\begin{equation}
    \mathbf{p}_{i}=\mathbf{x}_{i,t_0},
\end{equation}
where $t_0$ denotes the canonical frame. We then combine the motion direction and spatial position into a unified spatio-temporal feature:
\begin{equation}
    \mathbf{E}_{i}=\left[\lambda_s \{\mathbf{s}_{i,t}\}_{t=1}^{T-1}, \lambda_p \mathbf{p}_{i}\right]
\end{equation}
where $\lambda_s$ and $\lambda_p$ balance the contributions of motion direction and spatial position. In practice, $\mathbf{p}_{i}$ is normalized to have a comparable scale to the unit velocity-direction features.
    
\noindent
\textbf{Part Assignment and Initialization.}
The assignment of each Gaussian to a specific kinematic cluster $k \in \{0, \dots, K-1\}$ is determined as follows:
\begin{itemize}
    \item \textbf{Static Identification ($k=0$):} We compute the accumulated motion magnitude:
    \begin{equation}
        M_i=\sum_{t=1}^{T-1}\tilde{s}_{i,t}.
    \end{equation}
    If $M_i$ falls below a robust threshold $\tau$ (determined by the $20^{th}$ percentile of motion distribution across the scene), the Gaussian is classified as part of the static base.
    \item \textbf{Dynamic Assignment ($k > 0$):} For non-static Gaussians, we cluster them into $K-1$ groups using K-means on the spatio-temporal feature $\mathbf{E}_i$:
    \begin{equation}
        k_i=\arg\min_{k\in\{1,\dots,K-1\}}\left\|\mathbf{E}_{i}-\boldsymbol{\mu}_{k}\right\|_2^2+ 1
    \end{equation}
    where $\boldsymbol{\mu}_{k}$ denotes the $k$-th K-means cluster center in the spatio-temporal feature space.
    \item \textbf{Coefficient Initialization:} To facilitate the differentiable refinement in Stage 2, we initialize the distance distribution $\mathbf{d}_i$ for the soft-assignment logits. For dynamic clusters, the initial distance is proportional to the negative log-probability of the energy distribution: 
    \begin{equation}
        d_{i,k}=\left\|\mathbf{E}_{i}-\boldsymbol{\mu}_{k}\right\|_2^2.
    \end{equation}
\end{itemize}
Finally, we recompute the cluster centers $\mathbf{C}_k$ in the canonical frame using the spatial median of all Gaussians assigned to part $k$. This strategy ensures that even if the initial 3D tracks are noisy, the resulting part decomposition is grounded in the temporal logic of the object's kinematics.

\subsection{Mathematical Formulations of Articulated Kinematics}
\label{sec:supp_kinematics}
In the second stage of our pipeline, we transition from soft $\mathbb{SE}(3)$ motion bases to a rigorous kinematic parameterization to enforce physical consistency. This section provides the exhaustive mathematical derivation for the revolute and prismatic joint models utilized in Articulat3D.

\noindent
\textbf{Kinematic Parameters.} 
For each articulated object part $k \in \{1, ...,K\}$, the motion is governed by a specific joint model. We define the following learnable parameters:
\begin{itemize}
    \item \textbf{Joint Axis $\tilde{\mathbf{a}}_k \in \mathbb{R}^3$}: To ensure the rotation or translation axis remains unitlength during optimization, we derive the normalized axis $a_k$ as:
        \begin{equation}
            \mathbf{a}_k = \frac{\tilde{\mathbf{a}}_k}{\|\tilde{\mathbf{a}}_k\| + \varepsilon}
        \end{equation}
        where $\varepsilon = 10^{-7}$ is a small constant for numerical stability.
    \item \textbf{Pivot Point $\mathbf{c}_k \in \mathbb{R}^3$}: This represents the center of rotation for revolute joints or a point through which the translation axis passes for prismatic joints.
    \item \textbf{Joint Scalar $q_k(t) \in \mathbb{R}$}: A per-frame learnable value representing the rotation angle $\theta_k(t)$ (in radians) or the translational displacement $d_k(t)$.
\end{itemize}

\noindent
\textbf{Revolute Joint Model.} 
For rotational motion, we utilize the Rodrigues' rotation formula to compute the per-frame rotation matrix $\mathbf{R}_k(t) \in \mathbb{SO}(3)$ based on the joint scalar $\theta_k(t) = q_k(t)$. The matrix is defined as:
\begin{equation}
    \mathbf{R}_k(t) = \exp\left( \theta_k(t) [\mathbf{a}_k]_{\times} \right) 
    = \mathbf{I} + \sin(\theta_k(t)) [\mathbf{a}_k]_{\times} + (1 - \cos(\theta_k(t))) [\mathbf{a}_k]_{\times}^2
\end{equation}
where $[\mathbf{a}_k]_\times$ denotes the skew-symmetric matrix of the normalized axis vector $\mathbf{a}_k$. To account for the fact that the rotation occurs around a pivot $\mathbf{c}_k$ rather than the world origin, the corresponding translation vector $\mathbf{t}_k(t)$ is derived by enforcing the pivot as a fixed point under rotation:
\begin{equation}
    \mathbf{t}_k(t) = \mathbf{c}_k - \mathbf{R}_k(t)\mathbf{c}_k
\end{equation}
This ensures that the transformation $\mathbf{T}_k(t) = [\mathbf{R}_k(t), \mathbf{t}_k(t)]$ represents a pure rotation in the local coordinate system of the pivot.

\noindent
\textbf{Prismatic Joint Model.} 
For translational motion, the rotation matrix remains an identity matrix, and the movement is characterized as a displacement $d_k(t) = q_k(t)$ along the axis $\mathbf{a}_k$:
\begin{equation}
    \mathbf{R}_k(t) = \mathbf{I}, \quad \mathbf{t}_k(t) = d_k(t)\mathbf{a}_k
\end{equation}
This formulation ensures that all points within the Gaussian cluster $k$ undergo a strictly synchronized linear shift.

\subsection{Mathematical Formulations of Joint Initialization}
\label{sec:supp_jointinit}
To bridge the gap between the soft motion bases and the rigid kinematic constraints, we first establish a discrete part-level assignment. Leveraging the per-Gaussian motion coefficients $\mathbf{w}_i$, we perform a hard assignment for each Gaussian $i$ to a specific part $k \in \{1, \dots, K\}$ via $k_i^* = \arg\max_b w_{i}^{(b)}$.

Given the inherent noise in initial actual trajectories of gaussians, we employ a trimmed mean strategy to extract a stable center trajectory $c_k(t)$ for each dynamic part. At each frame $t$, we first determine the spatial median $\hat{c}_k(t) = \text{median}_{i \in S_k}(\mu_i(t))$. We then compute the Euclidean distances $d_i(t) = \|\mu_i(t) - \hat{c}_k(t)\|_2$ and retain only the $80\%$ subset of points $S_t$ with the smallest $d_i(t)$. The robust center trajectory is synthesized as:
\begin{equation}
    c_k(t) = \frac{1}{|S_t|}\sum_{i \in S_t} \mu_i(t)
\end{equation}

The synthesized trajectory $c_k(t)$ is analyzed via Principal Component Analysis (PCA) to determine joint types and parameters. By examining the principal components of the trajectory, we distinguish between revolute joints (forming planar arcs) and prismatic joints (exhibiting linear motion), yielding the initial axis direction $\mathbf{a}_k$ and pivot point $\mathbf{c}_k$.

Unlike the weighted interpolation utilized in Sec.~\ref{sec:prior_init}, every Gaussian $i$ assigned to part $k$ undergoes an identical global transformation $\mathbf{T}_k(t) = [\mathbf{R}_k(t), \mathbf{t}_k(t)] \in \mathbb{SE}(3)$. For any Gaussian within cluster $k$, its pose at time $t$ is strictly defined by applying the part-specific rigid transformation to its canonical state:
\begin{equation}
    \mu_i(t) = \mathbf{R}_k(t)\mu_i(0) + \mathbf{t}_k(t)
\end{equation}
\begin{equation}
    R_i(t) = \mathbf{R}_k(t)R_i(0)
\end{equation}
Through this geometric optimization parameterization, the motion of object parts is enforced as a strictly collective rigid-body transformation, which ensures that the spatial configuration of points within each component remains invariant over time.

\subsection{Detailed Formulation of Regularization Losses}
\label{sec:supp_loss}
To further stabilize the optimization in Stage 2 and ensure physical plausibility in the absence of multi-view supervision, we introduce two critical regularization terms: the Acceleration Loss ($\mathcal{L}_{acc}$) and the Depth-Stability Loss ($\mathcal{L}_{z}$).

\noindent
\textbf{Acceleration Loss ($\mathcal{L}_{acc}$)}
The Acceleration Loss is designed to enforce temporal smoothness by penalizing high-frequency oscillations in the articulated motion. Unlike unconstrained tracking methods that may produce jittery trajectories, we apply this loss directly to the learnable joint scalars $q_k(t)$ (representing either rotation angles $\theta_k(t)$ or translational displacements $d_k(t)$). The loss is formulated as the squared second-order finite difference of the motion sequence:
\begin{equation}
    \mathcal{L}_{\text{acc}} = \sum_{t,k} \| \ddot{q}_k(t) \|^2 = \sum_{t,k} \| q_k(t+1) - 2q_k(t) + q_k(t-1) \|^2,
\end{equation}
By minimizing the acceleration (the second derivative of position over time), $\mathcal{L}_{acc}$ encourages the articulated parts to move with a constant or smoothly changing velocity. This is particularly effective for suppressing the noise inherent in monocular tracking priors and ensuring that the reconstructed digital twin exhibits a natural, physically consistent evolutionary process.

\noindent
\textbf{Depth-Stability Loss ($\mathcal{L}_{z}$)}
In monocular reconstruction, a common artifact known as the "breathing effect" occurs when the optimization erroneously adjusts the depth of Gaussians (the $Z$-coordinate in camera space) to minimize color re-projection errors, even when no such physical motion exists. To mitigate this depth-scale ambiguity, we introduce the Depth-Stability Loss ($\mathcal{L}_{z}$).This loss penalizes the displacement of Gaussian centers along the camera’s optical axis ($Z$-axis) that cannot be explained by the explicit kinematic model. Let $\Delta \mathbf{x}_{i,t}^{(z)}$ be the $Z$-component of the predicted 3D displacement for Gaussian $i$ at time $t$. The loss is defined as:
\begin{equation}
    \mathcal{L}_{z} = \sum_{i=1}^{N} \sum_{t=1}^{T} \omega_i \cdot \| \Delta \mathbf{x}_{i,t}^{(z)} \|^2
\end{equation}
where $\omega_i$ is a confidence weight derived from the Stage 1 motion priors. By specifically constraining the $Z$-axis variance, $\mathcal{L}_{z}$ forces the model to prioritize explaining the visual observations through rotation and lateral translation—which are well-observed in the image plane—rather than through non-physical depth-wise "pulsing." This loss is essential for preserving the static canonical shape of the object during complex articulations.

\section{Experimental Details}
\subsection{Evaluation Protocol and Baseline Configurations}
For a fair comparison, RSRD~\cite{kerr2024robot} is adapted with NKSR mesh reconstruction, Articulate Anything~\cite{le2024articulate} is provided with the PartNet-Mobility library, and iTACO~\cite{peng2025itaco} is given ground-truth depth.

For quantitative results, N/A entries in Tab.~\ref{tab:exp} indicate inapplicable metrics: EPE is omitted for datasets lacking ground-truth trajectories; view synthesis metrics (PSNR/SSIM/LPIPS) are excluded for Articulate Anything~\cite{le2024articulate} and iTACO~\cite{peng2025itaco} as they do not support rendering. Notably, Shape of Motion~\cite{wang2025shape} is omitted from joint and part-level evaluations as it produces a holistic dynamic representation without explicit kinematic segmentation or joint parameterization. 

For qualitative results, we visualize reconstruction results as meshes, using grey for static components, various colors for dynamic parts, and red arrows for joint axes. Failed indicates cases where RSRD~\cite{kerr2024robot} failed to reconstruct geometry. Since baseline methods such as Articulate Anything~\cite{le2024articulate} and iTACO~\cite{peng2025itaco} only provide static meshes rather than per-frame poses, their results are shown in fixed configurations that may not match the GT articulated states.

\begin{table}[t]
 \centering
    \caption{Hyperparameter sensitivity analysis for \our. We evaluate the impact of varying learning rates for key parameters including motion coefficients, joint axes, and pivot points. The default settings (indicated in blue) consistently yield the optimal balance between convergence speed and reconstruction accuracy.}
    \vspace{-5pt}
    \label{tab:supp_hyper_ablation}
    \scriptsize
    \setlength{\tabcolsep}{2pt}
    \resizebox{\linewidth}{!}{
        \begin{tabular}{lcccccccc}
        \toprule
        \multirow{2}{*}{Parameter Configuration} & \multicolumn{2}{c}{Joint Estimation} & \multicolumn{3}{c}{Reconstruction} & \multicolumn{1}{c}{3D Tracking} & \multicolumn{2}{c}{View Synthesis} \\
        \cmidrule(lr){2-3} \cmidrule(lr){4-6} \cmidrule(lr){7-7} \cmidrule(lr){8-9}
        & Axis Err $\downarrow$ & Position Err $\downarrow$ & CD-w $\downarrow$ & CD-m $\downarrow$ & CD-s $\downarrow$ & EPE $\downarrow$ & PSNR $\uparrow$ & SSIM $\uparrow$ \\
        \midrule
        
        \multicolumn{9}{c}{\textbf{Motion Coefficients Learning Rate ($\eta_{w}$)}} \\
        $\eta_{w} = 1 \times 10^{-2}$ & 1.25 & 2.45 & 1.05 & 1.42 & 0.92 & 0.06 & 34.20 & 0.96 \\
        \rowcolor[HTML]{D7F6FF} $\eta_{w} = 5 \times 10^{-3}$ (Default) & \textbf{0.53} & \textbf{0.65} & \textbf{0.76} & \textbf{0.84} & \textbf{0.85} & \textbf{0.04} & \textbf{37.80} & \textbf{0.98} \\
        $\eta_{w} = 1 \times 10^{-3}$ & 0.89 & 1.12 & 0.95 & 1.10 & 0.88 & 0.05 & 36.15 & 0.97 \\
        \midrule
    
        \multicolumn{9}{c}{\textbf{Joint Axis \& Pivot Point Learning Rate ($\eta_{a}, \eta_{c}$)}} \\
        $\eta_{a,c} = 1 \times 10^{-3}$ & 2.14 & 5.82 & 2.30 & 3.45 & 1.25 & 0.08 & 31.50 & 0.94 \\
        \rowcolor[HTML]{D7F6FF} $\eta_{a,c} = 1 \times 10^{-4}$ (Default) & \textbf{0.53} & \textbf{0.65} & \textbf{0.76} & \textbf{0.84} & \textbf{0.85} & \textbf{0.04} & \textbf{37.80} & \textbf{0.98} \\
        $\eta_{a,c} = 5 \times 10^{-5}$ & 0.68 & 0.92 & 0.82 & 0.95 & 0.86 & 0.04 & 37.42 & 0.98 \\
        \midrule
    
        \multicolumn{9}{c}{\textbf{Joint Scalar Learning Rate ($\eta_{q}$)}} \\
        $\eta_{q} = 1 \times 10^{-2}$ & 0.94 & 1.56 & 1.20 & 1.65 & 0.98 & 0.07 & 35.80 & 0.97 \\
        \rowcolor[HTML]{D7F6FF} $\eta_{q} = 5 \times 10^{-3}$ (Default) & \textbf{0.53} & \textbf{0.65} & \textbf{0.76} & \textbf{0.84} & \textbf{0.85} & \textbf{0.04} & \textbf{37.80} & \textbf{0.98} \\
        \bottomrule
        \end{tabular}
    }
\end{table}

\subsection{Hyperparameter Settings}
We provide a comprehensive sensitivity analysis of the core learning rates governing our multi-stage optimization pipeline in Tab. \ref{tab:supp_hyper_ablation}. Specifically, we evaluate the impact of varying the learning rates for the motion coefficients ($\eta_w$), the kinematic joint parameters ($\eta_{a,c}$), and the per-frame joint scalars ($\eta_q$). The results demonstrate that Articulat3D is relatively robust to moderate variations in these hyperparameters, yet certain thresholds are critical for maintaining physical consistency. For instance, excessively high learning rates for the joint axis ($\eta_{a}$) can lead to instability in the non-convex $SO(3)$ optimization space, whereas insufficient learning rates for the motion coefficients ($\eta_w$) hinder the clear emergence of part boundaries during the first stage. Our empirically determined default configurations (highlighted in blue) consistently yield the optimal trade-off between kinematic accuracy and surface reconstruction fidelity across both synthetic and real-world benchmarks.

\begin{table}[t]
    \centering
    \caption{Runtime comparison across different datasets. We report the average time (in minutes) per scene. All experiments were conducted on a single NVIDIA RTX 4090 GPU.}
    \label{tab:runtime}
    \scriptsize
    \setlength{\tabcolsep}{2pt} 
    \resizebox{0.6\linewidth}{!}{
        \begin{tabular}{lcc}
        \toprule
        \multirow{2}{*}{Method} & \multicolumn{2}{c}{Datasets (min)} \\
        \cmidrule(lr){2-3}
        & Video2Articulation & Articulat3D-Sim \\
        \midrule
        Shape-of-Motion~\cite{wang2025shape} & $\sim$63.28 & $\sim$85.77 \\
        RSRD~\cite{kerr2024robot} & $\sim$43.51 & $\sim$67.93 \\
        iTACO~\cite{peng2025itaco} & \textbf{$\sim$10.77} & \textbf{$\sim$20.58} \\
        \midrule
        \rowcolor[HTML]{D7F6FF} \textbf{}{\our (Ours)} & \underline{$\sim$34.52} & \underline{$\sim$58.27} \\
        \bottomrule
        \end{tabular}
    }
    \vspace{2pt}
    \begin{flushleft}
    \scriptsize *Note: Articulate Anything~\cite{le2024articulate} is excluded from this comparison as it retrieves models from an existing library rather than performing a per-scene reconstruction process. 
    \end{flushleft}
\end{table}

\subsection{Training Efficiency}
Tab.~\ref{tab:runtime} summarizes the training time for all evaluated methods on a single NVIDIA RTX 4090 GPU. While Articulat3D requires a moderate optimization period compared to iTACO~\cite{peng2025itaco}, this marginal increase in computational cost enables a substantial leap in reconstruction fidelity, joint estimation accuracy, and physical consistency. Notably, our framework remains significantly more efficient than prior rendering-based optimization methods such as RSRD~\cite{kerr2024robot} and Shape-of-Motion~\cite{wang2025shape}, which often require over an hour per scene, demonstrating an optimal trade-off between speed and performance.

\begin{table}[t]
    \centering
    \scriptsize
    \caption{\textbf{Quantitative comparisons} on three datasets}
    \label{tab:supp_robust}
    \setlength{\tabcolsep}{1pt}
    \resizebox{\linewidth}{!}{
        \begin{tabular}{l|cc|ccc|c|ccc}
        \toprule
        Method & Axis Err $\downarrow$ & Pos Err $\downarrow$ & CD-w $\downarrow$ & CD-m $\downarrow$ & CD-s $\downarrow$ & EPE $\downarrow$ & PSNR $\uparrow$ & SSIM $\uparrow$ & LPIPS $\downarrow$ \\
        
        \midrule
        \multicolumn{10}{c}{\textbf{Video2Articulation-S}} \\ 
        \rowcolor[HTML]{D7F6FF} \our & 1.60{\tiny±1.82} & 1.83{\tiny±1.17} & 0.82{\tiny±0.94} & 1.12{\tiny±1.53} & 1.82{\tiny±0.65} & N/A & 35.91{\tiny±2.36} & 0.98{\tiny±0.01} & 0.05{\tiny±0.02} \\
        
        \midrule
        \multicolumn{10}{c}{\textbf{\our-Sim}} \\ 
        \rowcolor{gray!10} noise mask & 0.54{\tiny±0.81} & 0.63{\tiny±0.93} & 0.79{\tiny±1.42} & 0.90{\tiny±1.14} & 0.86{\tiny±0.67} & 0.05{\tiny±0.07} & 37.45{\tiny±2.11} & 0.97{\tiny±0.01} & 0.03{\tiny±0.01}  \\
        noise track & 0.56{\tiny±0.89}	& 0.67{\tiny±0.99}	& 0.78{\tiny±1.06}	& 0.88{\tiny±1.24} & 0.86{\tiny±0.69}	& 0.05{\tiny±0.08}	& 37.62{\tiny±2.39}	& 0.98{\tiny±0.02} & 0.03{\tiny±0.01}  \\
        
        \rowcolor[HTML]{D7F6FF} \our & 0.53{\tiny±0.70} & 0.65{\tiny±0.82} & 0.76{\tiny±1.52} & 0.84{\tiny±1.12} & 0.85{\tiny±0.62} & 0.04{\tiny±0.06} & 37.80{\tiny±2.04} & 0.98{\tiny±0.01} & 0.03{\tiny±0.01}  \\
        
        \midrule
        \multicolumn{10}{c}{\textbf{\our-Real}} \\ 
        \rowcolor[HTML]{D7F6FF} \our & 2.02{\tiny±3.86} & 1.61{\tiny±2.44} & N/A  & N/A  & N/A  & N/A & 26.73{\tiny±2.39} & 0.93{\tiny±0.02} & 0.14{\tiny±0.01} \\
        
        \midrule
        \multicolumn{10}{c}{\textbf{VideoArtGS-20}} \\ 
        \rowcolor{gray!10} VideoArtGS & 0.34{\tiny±0.80} & 0.10{\tiny±0.10} & \textbf{0.09{\tiny±0.09}} & \textbf{0.26{\tiny±0.61}} & 0.24{\tiny±0.58} &  N/A & 37.42{\tiny±2.36} & 0.97{\tiny±0.01} & 0.02{\tiny±0.01}  \\
        \rowcolor[HTML]{D7F6FF} \our & \textbf{0.27{\tiny±0.79}} & \textbf{0.09{\tiny±0.12}} & 0.15{\tiny±0.22}  & 0.27{\tiny±0.76}  & \textbf{0.22{\tiny±0.46}}  & N/A & \textbf{38.84{\tiny±4.61}} & \textbf{0.98{\tiny±0.02}} & \textbf{0.01{\tiny±0.01}} \\ 
        
        \bottomrule
        \end{tabular}
    }
\end{table}

\subsection{Extended Experiments}
\noindent
\textbf{Robustness of Metric Values.}
To better characterize the robustness and variability of Articulat3D, we report all quantitative results as the mean ± standard deviation in Tab.~\ref{tab:supp_robust}.


\noindent
\textbf{Robustness to upstream prediction errors.}
We further evaluate the sensitivity of Articulat3D to errors introduced by its upstream segmentation and tracking components. For segmentation, we apply region-level perturbations to the SAM3 masks, since practical segmentation errors are generally spatially correlated rather than independent at the pixel level.

Specifically, we randomly insert one or two spatially contiguous false-positive foreground regions into the background of each frame. Their shapes are sampled from ellipses, rectangles, and irregular polygons, and their areas are sampled relative to the image size. We then rerun the complete pipeline with the perturbed masks. As shown in Tab.~\ref{tab:supp_robust} (rows noise mask and noise track), the resulting Chamfer Distance and joint-axis metrics change only marginally. This robustness benefits from the dense multi-view observations: local errors appearing in only a few views lack consistent geometric support in 3D and therefore have limited influence on the final reconstruction.

To assess robustness to tracking errors, we perturb the TAPIP3D trajectories with zero-mean Gaussian noise. Given a predicted trajectory point $\mathbf{x}*{i,t}\in\mathbb{R}^{3}$, we construct
\begin{equation}
    \tilde{\mathbf{x}}_{i,t} = \mathbf{x}_{i,t} + \boldsymbol{\epsilon}_{i,t}, \quad \boldsymbol{\epsilon}_{i,t} \sim \mathcal{N} (0, \sigma^2 \mathbf{I})
\end{equation}
where $\sigma$ controls the noise magnitude. The perturbation is applied only to the 3D coordinates, while visibility, confidence, temporal correspondences, and static-point labels remain unchanged. The results remain stable under moderate perturbations, indicating that Articulat3D does not require perfectly accurate TAPIP3D trajectories. This is because the trajectories primarily initialize the dual-quaternion motion bases, joint axes, and joint-type inference, rather than serving as hard ground-truth constraints throughout optimization. Subsequent multi-view appearance, depth, and geometric-consistency objectives can therefore compensate for moderate errors in the initial trajectories.

\noindent
\textbf{Evaluation on real-world sequences.}
Since accurate ground-truth geometry and articulation parameters are difficult to acquire for real captured objects, our original evaluation on Articulat3D-Real mainly relied on image-level rendering metrics. To enable a more direct assessment of articulation estimation, we manually annotate the joint-axis directions and positions for a subset of the real-world sequences. We then compute the axis direction error and axis position error against these annotations. As reported in Tab.~\ref{tab:supp_robust} (Row 5), these additional metrics complement PSNR, SSIM, and LPIPS by directly evaluating the accuracy of the recovered joint parameters in real-world settings.


\noindent
\textbf{Evaluation against VideoArtGS.}
We further compare Articulat3D with the recent VideoArtGS~\cite{liu2025videoartgs} baseline on the VideoArtGS-20 benchmark, following its original evaluation setting. As shown in Tab.~\ref{tab:supp_robust} (Rows 6-7), Articulat3D achieves competitive or better performance across the evaluated metrics. In particular, our method improves joint-axes estimation and novel-view synthesis quality, while maintaining comparable reconstruction accuracy. These results demonstrate that Articulat3D generalizes effectively to an external benchmark and remains competitive with a strong recent method under its native evaluation protocol.

\section{Dataset: Articulat3D-Sim}
\label{sec:supp_dataset}
\begin{figure}
    \centering
    \includegraphics[width=0.8\linewidth]{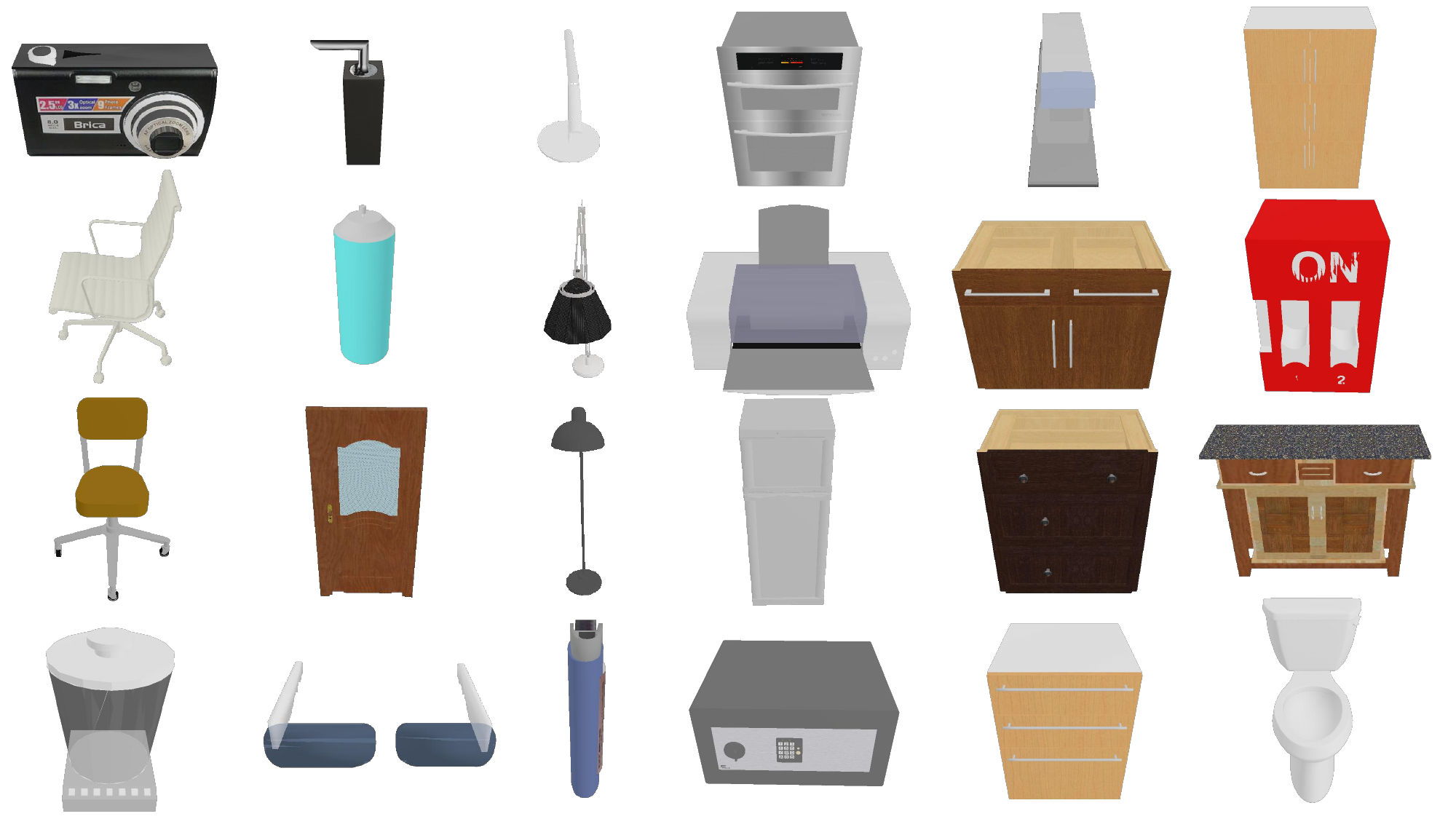}
    \caption{Visualization of our Articulat3D-Sim dataset.}
    \label{fig:supp_data}
\end{figure}

We introduce and evaluate our method on Articulat3D-Sim, a newly curated dataset comprising 17 object categories from PartNet-Mobility~\cite{xiang2020sapien}: Camera, Chair, Coffee Machine, Dispenser, Door, Eyeglasses, Lamp, Lighter, Oven, Printer, Refrigerator, Safe, Stapler, Storage Furniture, Switch, Table, and Toilet. This dataset consists of 24 sequences featuring complex kinematics with 2 to 7 movable parts per object. To simulate realistic human-object interaction scenarios, each trajectory is meticulously designed to initiate with synchronized part articulation and large-scale camera movement, concluding with a full 360-degree orbit around the object for a holistic inspection. This setup provides a challenging benchmark that requires the model to decouple intricate geometry and motion under significant viewpoint changes. Further details and visualizations are available in Fig.~\ref{fig:supp_data}.

\end{document}